\documentclass[11pt,a4paper]{article}
\usepackage[hyperref]{emnlp2020}
\usepackage{times}
\usepackage{latexsym}

\usepackage{cancel}
\usepackage{microtype}
\usepackage[algo2e]{algorithm2e} 
\aclfinalcopy 
\usepackage{algorithm}
\usepackage{algorithmic}

\usepackage{amssymb}
\usepackage{amsthm}
\newtheorem{theorem}{Theorem}[section]

\usepackage{graphicx}
\usepackage{amsmath}
\usepackage{amssymb}
\usepackage{multirow}
\usepackage{makecell}
\usepackage{url}

\usepackage{subcaption}
\usepackage{verbatim}
\usepackage{url}
\usepackage{array}
\usepackage{balance}
\usepackage{color}
\usepackage{soul}
\usepackage{hyperref}
\newcolumntype{L}[1]{>{\raggedright\let\newline\\\arraybackslash\hspace{0pt}}m{#1}}
\setul{2pt}{2pt}

\usepackage{pifont}

\renewcommand{\vec}[1]{\mathbf{#1}}

\DeclareMathOperator*{\argmax}{arg\,max}

\usepackage{changepage}
\usepackage{booktabs, tabularx}
\usepackage{arydshln}

\usepackage{pgfplots}
\usepackage{breakurl}
\usepackage{times}
\usepackage{latexsym}
\hypersetup{colorlinks}
\usepackage{booktabs}

\usepackage{url}
\newcounter{Lcount}
\newcommand{\squishenum}{
\begin{list}{\arabic{Lcount}. }
{ \usecounter{Lcount}
\setlength{\itemsep}{0pt}
\setlength{\parsep}{0pt}
\setlength{\topsep}{0pt}
\setlength{\partopsep}{0pt}
\setlength{\leftmargin}{2em}
\setlength{\labelwidth}{1.5em}
\setlength{\labelsep}{0.5em} } }

\newcommand{\RNum}[1]{\uppercase\expandafter{\romannumeral #1\relax}}
\newcommand{\squishletter}{
\begin{list}{\alph{Lcount}. }
{ \usecounter{Lcount}
\setlength{\itemsep}{0pt}
\setlength{\parsep}{0pt}
\setlength{\topsep}{0pt}
\setlength{\partopsep}{0pt}
\setlength{\leftmargin}{2em}
\setlength{\labelwidth}{1.5em}
\setlength{\labelsep}{0.5em} } }

\newcommand{\squishlist}{
\begin{list}{$\bullet$}
{ \usecounter{Lcount}
\setlength{\itemsep}{0pt}
\setlength{\parsep}{0pt}
\setlength{\topsep}{0pt}
\setlength{\partopsep}{0pt}
\setlength{\leftmargin}{2em}
\setlength{\labelwidth}{1.5em}
\setlength{\labelsep}{0.5em} } }

\newcommand{\squishend}{
\end{list} }

\usepackage[export]{adjustbox}
\usepackage{tikz}
\usepackage{pgf}
\usepackage{tikz-qtree}
\usepackage{xcolor}
\usepackage{multirow}
\usetikzlibrary{arrows,decorations.pathmorphing,backgrounds,positioning,fit,petri,shapes.misc, arrows.meta,shapes.geometric,decorations.markings,calc,shadows.blur,decorations.pathreplacing,quotes,matrix,shapes.symbols,patterns}
\definecolor{fontgray}{RGB}{44, 62, 80}
\definecolor{myred}{RGB}{235, 47, 6} 
\definecolor{summertime}{RGB}{245, 205, 121}
\definecolor{darkgrass}{RGB}{0, 148, 50}
\definecolor{myblue}{RGB}{0, 168, 255}
\definecolor{mygray}{RGB}{158, 158, 158}
\definecolor{puffin}{RGB}{250, 152, 58}
\definecolor{lowpurple}{RGB}{210, 180, 222}
\definecolor{lowblue}{RGB}{102,178,255}
\definecolor{lowred}{RGB}{245, 183, 177}

\newcommand{\pluseq}{\mathrel{+}=}

\newcommand*{\affaddr}[1]{#1} 
\newcommand*{\affmark}[1][*]{\textsuperscript{#1}}

\title{Position-Aware Tagging for Aspect Sentiment Triplet Extraction}

\author{%
Lu Xu\affmark[* 1, 2]\thanks{$*$ Equal contribution.  Lu Xu is under the Joint PhD Program between Alibaba and Singapore University of Technology and Design. The work was done when Hao Li was a PhD student in Singapore University of Technology and Design.},
\thanks {Accepted as a long paper in EMNLP 2020 (Conference on Empirical Methods in Natural Language Processing).} 
Hao Li\affmark[* 1, 3], Wei Lu\affmark[1], Lidong Bing\affmark[2]\\
\affaddr{\affmark[1] StatNLP Research Group, Singapore University of Technology and Design}\\
\affaddr{\affmark[2] DAMO Academy, Alibaba Group}~~ \affaddr{\affmark[3]ByteDance}\\
\tt{xu\_lu@mymail.sutd.edu.sg, hao.li@bytedance.com}\\
\tt{ luwei@sutd.edu.sg, l.bing@alibaba-inc.com}\\
}
\renewcommand\footnotemark{}
\date{}

\pgfplotsset{
    /pgfplots/area legend/.style={
        legend image code/.code={%
        \draw[#1] (0cm,-0.1cm) rectangle (0.1cm,0.15cm);
    }
    }
}

\begin{document}
\maketitle
\begin{abstract}
Aspect Sentiment Triplet Extraction (ASTE) is the task of extracting the triplets of target entities, their associated sentiment, and opinion spans explaining the reason for the sentiment.
Existing research efforts mostly solve this problem using pipeline approaches, which break the triplet extraction process into several stages. 
Our observation is that the three elements within a triplet are highly related to each other, and this motivates us to build a joint model to extract such triplets using a sequence tagging approach.
However, how to effectively design a tagging approach to extract the triplets that can capture the rich interactions among the elements is a challenging research question.
In this work, we propose the first end-to-end model with a novel {\em position-aware} tagging scheme that is capable of jointly extracting the triplets.
Our experimental results on several existing datasets show that jointly capturing elements in the triplet using our approach leads to improved performance over the existing approaches.
We also conducted extensive experiments to investigate the model effectiveness and robustness\footnote{We release our code at \url{https://github.com/xuuuluuu/Position-Aware-Tagging-for-ASTE}}.

\end{abstract}


\section{Introduction}
\label{sec:intro}

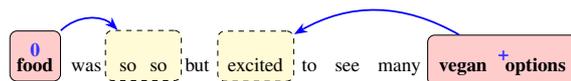
\begin{figure}[t!]
		\centering
		\adjustbox{max width=1.0\linewidth}{
			\begin{tikzpicture}[node distance=1.0mm and 1.0mm, >=Stealth, 
			wordnode/.style={draw=none, minimum height=5mm, inner sep=0pt},
			chainLine/.style={line width=1pt,->, color=blue},
			opinionbox/.style={draw=black, rounded corners, fill=yellow!20, dashed},
			targetbox/.style={draw=black, rounded corners, fill=red!20}
			]

			\matrix (sent1) [matrix of nodes, nodes in empty cells, execute at empty cell=\node{\strut};]
			{
				\textbf{food} & [1mm] was &[1mm] so & so & [1mm] but &  [1mm]  excited & [1mm] to & [1mm] see & [1mm] many & [1mm] \textbf{vegan} & [1mm] \textbf{options}\\
			};
						
			\begin{pgfonlayer}{background}
			
			\node [targetbox, above=of sent1-1-1, yshift=-7mm, text height=-2mm, minimum height = 10mm, minimum width=10mm] (e1)  [] {\color{blue!80}\textbf{\textsc{0}}};	
			\node [opinionbox, above=of sent1-1-3, xshift=3mm, yshift=-6mm, text height=-5mm, minimum height = 10mm, minimum width=15mm] (o1)  [] {\color{blue!80}\textbf{\textsc{}}};
			
			\draw[chainLine,blue]   (e1) to[out=45,in=135] (o1);

			\node [targetbox, above=of sent1-1-10, xshift=8mm, yshift=-7mm, text height=-2mm, minimum height = 10mm, minimum width=30mm] (e2)  [] {\color{blue!80}\textbf{\textsc{+}}};			
			\node [opinionbox, above=of sent1-1-6, xshift=0mm, yshift=-7mm, text height=-5mm, minimum height = 10mm, minimum width=15mm] (o2)  [] {\color{blue!80}\textbf{\textsc{}}};
			
			\draw[chainLine,blue]   (e2) to[out=150,in=30] (o2);
				
			\end{pgfonlayer}

			\end{tikzpicture} 
		}
		\caption{ASTE with targets in bold in solid squares, their associated sentiment on top, and opinion spans in dashed boxes. The arc indicates connection between a target and the corresponding opinion span.}
		\vspace{-2mm}
		\label{fig:example}
\end{figure}

Designing effective algorithms that are capable of automatically performing sentiment analysis and opinion mining is a challenging and important task in the field of natural language processing~\cite{pang2008opinion,liu2010sentiment,ortigosa2014sentiment,smailovic2013predictive,li2010using}.
Recently, Aspect-based Sentiment Analysis ~\cite{pontiki-EtAl:2014:SemEval} or Targeted Sentiment Analysis~\cite{mitchell2013open} which focuses on extracting target phrases as well as the sentiment associated with each target, has been receiving much attention.
In this work, we focus on a relatively new task -- Aspect Sentiment Triplet Extraction (ASTE) proposed by~\citet{peng2019knowing}. 
Such a task is required to extract not only the targets and the sentiment mentioned above, but also the corresponding opinion spans expressing the sentiment for each target.
Such three elements: a target, its sentiment and the corresponding opinion span, form a triplet to be extracted.
Figure~\ref{fig:example} presents an example sentence containing two targets in solid boxes.
Each target is associated with a sentiment, where we use $+$ to denote the positive polarity, $0$ for neutral, and $-$ for negative.
Two opinion spans in dashed boxes are connected to their targets by arcs.
Such opinion spans are important, since they largely explain the sentiment polarities for the corresponding targets~\cite{qiu-etal-2011-opinion,yang-cardie-2012-extracting}.

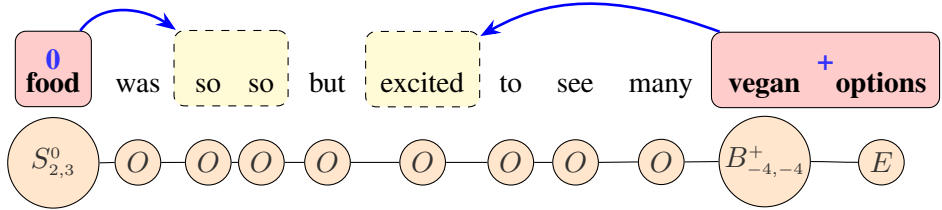
\begin{figure*}[t!]
		\centering
			\adjustbox{max width=1.0\linewidth}{
			\begin{tikzpicture}[node distance=1.0mm and 1.0mm, >=Stealth, 
			wordnode/.style={draw=none, minimum height=5mm, inner sep=0pt},
			chainLine/.style={line width=1pt,->, color=blue},
			opinionbox/.style={draw=black, rounded corners, fill=yellow!20, dashed},
			targetbox/.style={draw=black, rounded corners, fill=red!20},
			BStag/.style={shape=circle, draw=black, rounded corners, fill=orange!20, minimum height=12mm, inner sep=0pt},
			MEOtag/.style={shape=circle, draw=black, rounded corners, fill=orange!20, minimum height=6mm, inner sep=0pt}
			]

			\matrix (sent1) [matrix of nodes, nodes in empty cells, execute at empty cell=\node{\strut};]
			{
				\textbf{food} & [2mm] was &[2mm] so & [1mm]  so & [2mm] but &  [2mm]  excited & [2mm] to & [2mm] see & [2mm] many & [2mm] \textbf{vegan} & [2mm] \textbf{options}\\
			};
			
			
			\foreach \pos in {1}
				\node [BStag, below=of sent1-1-\pos, yshift=-0.9mm] (tag\pos)  [] {\color{black!80}\textsc{$S^{\scriptscriptstyle 0}_{\scriptscriptstyle 2,3}$}};

			\foreach \pos/\yoffset in {{2/-3.85},{3/-3.85},{4/-3.85},{5/-3.85},{6/-3.85},{7/-3.85},{8/-3.85},{9/-3}}
				\node [MEOtag, below=of sent1-1-\pos, yshift=\yoffset mm] (tag\pos)  [] {\color{black!80}\textsc{$O$}};
			
			\node [BStag, below=of sent1-1-10, yshift=0mm] (tag10)  [] {\color{black!80}\textsc{$B^{\scriptscriptstyle +}_{\scriptscriptstyle -4,-4}$}};
			
			\node [MEOtag, below=of sent1-1-11, yshift=-3.1mm] (tag11)  [] {\color{black!80}\textsc{$E$}};

			\foreach \pos/\npos in {{1/2},{2/3},{3/4},{4/5},{5/6},{6/7},{7/8},{8/9},{9/10},{10/11}}
				\draw[-]  (tag\pos) to[out=0,in=180] (tag\npos);

			\begin{pgfonlayer}{background}
			
			\node [targetbox, above=of sent1-1-1, yshift=-7mm, text height=-2mm, minimum height = 10mm, minimum width=10mm] (e1)  [] {\color{blue!80}\textbf{\textsc{0}}};	
			\node [opinionbox, above=of sent1-1-3, xshift=3mm, yshift=-6mm, text height=-5mm, minimum height = 10mm, minimum width=15mm] (o1)  [] {\color{blue!80}\textbf{\textsc{}}};
			
			\draw[chainLine,->]   (e1) to[out=55,in=145] (o1);

			\node [targetbox, above=of sent1-1-10, xshift=8mm, yshift=-6.3mm, text height=-2mm, minimum height = 10mm, minimum width=30mm] (e2)  [] {\color{blue!80}\textbf{\textsc{+}}};			
			\node [opinionbox, above=of sent1-1-6, xshift=0mm, yshift=-7mm, text height=-5mm, minimum height = 10mm, minimum width=15mm] (o2)  [] {\color{blue!80}\textbf{\textsc{}}};		
			
			\draw[chainLine,->]  (e2) to[out=160,in=30] (o2);
				
			\end{pgfonlayer}

			\end{tikzpicture} 
		}		
		\caption{The position-aware tagging scheme for the example instance.}
		\label{fig:structure}
	\end{figure*}

This ASTE problem was basically untouched before, and the only existing work that we are aware of~\cite{peng2019knowing} employs a 2-stage pipeline approach.
At the first stage, they employ a unified tagging scheme which fuses the target tag based on the $BIOES$\footnote{$BIOES$ is a common tagging scheme for sequence labeling tasks, and $BIOES$ denotes ``begin, inside, outside, end and single'' respectively.} tagging scheme, and sentiment tag together.
Under such a unified tagging scheme, they proposed methods based on Long Short-Term Memory networks (LSTM) \cite{lstm97}, Conditional Random Fields (CRF) \cite{lafferty2001conditional} and Graph Convolutional Networks (GCN) \cite{kipf2017semi} to perform sequence labeling to extract targets with sentiment as well as opinion spans.
At the second stage, they use a classifier based on Multi-Layer Perceptron (MLP) to pair each target (containing a sentiment label) with the corresponding opinion span to obtain all the valid triplets.

One important observation is that the three elements in a triplet are highly related to each other. 
Specifically, sentiment polarity is largely determined by an opinion span as well as the target and its context, and an opinion span also depends on the target phrase in terms of wording (e.g., an opinion span \textit{``fresh''} usually describes food targets instead of service).
Such an observation implies that jointly capturing the rich interaction among three elements in a triplet might be a more effective approach.
However, the $BIOES$ tagging scheme on which the existing approaches based comes with a severe limitation for this task:
such a tagging scheme without encoding any positional information fails to specify the connection between a target and its opinion span as well as the rich interactions among the three elements due to the limited expressiveness.
Specifically,  $BIOES$ uses the tag $B$ or $S$ to represent the beginning of a target. For example, in the example sentence in Figure \ref{fig:example}, \textit{``vegan''} should be labeled with $B$, but the tagging scheme does not contain any information to specify the position of its corresponding opinion \textit{``excited''}.
Using such a tagging scheme inevitably leads to an additional step to connect each target with an opinion span as the second stage in the pipeline approach.
{\color{black}The skip-chain sequence models~\cite{sutton2004collective, galley-2006-skip} are able to capture interactions between given input tokens which can be far away from each other.
However, they are not suitable for the ASTE task where the positions of targets and opinion spans are not explicitly provided but need to be learned.}


Motivated by the above observations, we present a novel approach that is capable of predicting the triplets jointly for ASTE. 
Specifically, we make the following contributions in this work:
\squishlist
\item  We present a novel position-aware tagging scheme that is capable of specifying the structural information for a triplet -- the connection among the three elements by enriching the label semantics with more expressiveness, to address the above limitation.
\item We propose a novel approach, \textbf{JET}, to \underline{J}ointly \underline{E}xtract the \underline{T}riplets based on our novel position-aware tagging scheme. 
Such an approach is capable of better capturing interactions among elements in a triplet by computing factorized features for the structural information in the ASTE task.
\item Through extensive experiments, the results show that our joint approach \textbf{JET} outperforms baselines significantly.
\squishend

\section{Our Approach}
Our objective is to design a model \textbf{JET} to extract the triplet of Target, Target Sentiment, and Opinion Span jointly.
We first introduce the new position-aware tagging scheme, followed by the model architecture. 
We next present our simple LSTM-based neural architecture for learning feature representations, followed by our method to calculate factorized feature scores based on such feature representations for better capturing the interactions among elements in a triplet.
Finally, we also discuss a variant of our model.


\subsection{Position-Aware Tagging Scheme}
To address the limitations mentioned above, we propose our position-aware tagging scheme by enriching expressiveness to incorporate position information for a target and the corresponding opinion span.
Specifically, we extend the tag $B$ and tag $S$ in the $BIOES$ tagging scheme to new tags respectively:
\setlength{\abovedisplayskip}{0pt} \setlength{\abovedisplayshortskip}{0pt}
\setlength{\belowdisplayskip}{4pt} \setlength{\belowdisplayshortskip}{4pt}
$$B^{\epsilon}_{j,k}, S^{\epsilon}_{j,k}$$
where $B^{\epsilon}_{j,k}$ with the sub-tag\footnote{We define the sub-tags of $B^{\epsilon}_{j,k}, S^{\epsilon}_{j,k}$ as $B$ and $S$ respectively, and the sub-tags of $I,O,E$ as themselves.} $B$ still denotes the beginning of a target, and $S^{\epsilon}_{j,k}$ with the sub-tag $S$ denotes a single-word target.
Note that $\epsilon \in \{+, 0, -\}$ denotes the sentiment polarity for the target, and $j,k$  indicate the position information which are  the distances 
between the two ends of an opinion span and the starting position of a target respectively.
Here, we use the term ``{\em offset}'' to denote such  position information for convenience.
We keep the other tags $I$, $E$, $O$ as is.
In a word, we use sub-tags $BIOES$ for encoding targets, $\epsilon$ for sentiment, and offsets for opinion spans under the new position-aware tagging scheme for the structural information.

For the example in Figure~\ref{fig:example}, under the proposed tagging  scheme, the tagging result is given in Figure~\ref{fig:structure}.
The single-word target \textit{``food''} is tagged with $S^{0}_{2,3}$, implying the sentiment polarity for this target is neutral (0). 
Furthermore, the positive offsets $2,3$ indicate that its opinion span is on the right and has distances of 2 and 3 measured at the left and right ends respectively, (i.e., \textit{``so so''}).
The second target is \textit{``vegan options''} with its first word tagged with $B^{+}_{-4,-4}$ and the last word tagged with $E$, implying the sentiment polarity is positive ($+$).
Furthermore, the negative offsets $-4,-4$ indicate that the opinion span \textit{``excited''} appears on the left of the target, and has distances of 4 and 4 measured at the left and right ends respectively, (i.e., \textit{``vegan''}).

\textcolor{black}{
Our proposed position-aware tagging scheme has the following theoretical property:
\begin{theorem} There is a one-to-one correspondence between a tag sequence and a combination of aspect sentiment triplets within the sentence as long as the targets do not overlap with one another, and each has one corresponding opinion span.\footnote{See the Appendix for detailed statistics on how often this condition is satisfied.}
\end{theorem} 
\begin{proof}
For a given triplet, we can use the following process to construct the tag sequence.
    First note that the sub-tags of our proposed tags $B^{\epsilon}_{j,k}, I, O, E, S^{\epsilon}_{j,k}$, are $B, I, O ,E, S$. The standard $BIOES$ tagset is capable of extracting all possible targets when they do not overlap with one another. 
    Next, for each specified target, the position information $j, k$ that specifies the position of its corresponding opinion span can be attached to the $B$ (or $S$) tag, resulting in $B_{j,k}$ (or $S_{j,k}$). Note that the opinion span can be any span within the sentence when $j, k$ are not constrained. Finally, we assign each extracted target its sentiment polarity $\epsilon$ by attaching it to the tag $B$ (or $S$), resulting in $B^\epsilon_{j,k}$ (or $S^\epsilon_{j,k}$). This construction process is unique for each combination of triplets. Similarly, given a tag sequence, we can reverse the above process to recover the combination of triplets.
\end{proof}
We would like to highlight that our proposed position-aware tagging scheme is capable of handling some special cases where the previous approach is unable to. For example, in the sentence \textit{``The salad is cheap with fresh salmon"}, there are two triplets, (\textit{``salad"}, \textit{``cheap with fresh salmon"}, positive)\footnote{We use the format (target, opinion spans, sentiment).} and (\textit{``salmon"}, \textit{``fresh"}, positive). The previous approach such as \cite{peng2019knowing}, which was based on a different tagging scheme, will not be able to handle such a case where the two opinion spans overlap with one another.}

\subsection{Our JET Model}

We design our novel \textbf{JET} model with CRF~\cite{lafferty2001conditional} and Semi-Markov CRF~\cite{sarawagi2004semi} based on our position-aware tagging scheme.
Such a model is capable of encoding and factorizing both token-level features for targets and segment-level features for opinion spans.

Given a sentence $\vec{x}$ with length $n$, we aim to produce the desired output sequence $\vec{y}$ based on the  position-aware  tagging scheme.
The probability of $\vec{y}$ is defined as:
\setlength{\abovedisplayskip}{4pt} \setlength{\abovedisplayshortskip}{4pt}
\setlength{\belowdisplayskip}{4pt} \setlength{\belowdisplayshortskip}{4pt}
\begin{eqnarray}
p(\vec{y}|\vec{x}) = \frac{\exp{(s(\vec{x},\vec{y}))} }{\sum_{\vec{y}' \in \vec{Y}_{\vec{x},M}}\exp(s(\vec{x},\vec{y}^{'}) )}
\end{eqnarray}
where $s(\vec{x},\vec{y})$ is a score function defined over the sentence $\vec{x}$ and the output structure $\vec{y}$, and $\vec{Y}_{\vec{x}, M}$ represents all the possible sequences under our position-aware tagging scheme with the offset constraint $M$, indicating the maximum absolute value of an offset.
The score $s(\vec{x},\vec{y})$ is defined as:
\setlength{\abovedisplayskip}{4pt} \setlength{\abovedisplayshortskip}{4pt}
\setlength{\belowdisplayskip}{4pt} \setlength{\belowdisplayshortskip}{4pt}
\begin{equation}
    \label{eq:score}
    s(\vec{x},\vec{y}) = \sum_{i=0}^{n} \psi_{\bar{\vec{y}}_i, \bar{\vec{y}}_{i+1}} + \sum_{i=1}^{n} \Phi_{\vec{y}_i}(\vec{x},i)
\end{equation}
where $\bar{\vec{y}}_i \in \{B,I,O,E,S\}$ returns the sub-tag of $\vec{y}_i$, $\psi_{\bar{\vec{y}}_i, \bar{\vec{y}}_{i+1}}$ represents the transition score: the weight of a ``transition feature'' -- a feature defined over two adjacent sub-tags $\bar{\vec{y}}_i$ and $\bar{\vec{y}}_{i+1}$, and $\Phi_{\vec{y}_i}(\vec{x},i)$ represents the factorized feature score with tag $\vec{y}_i$ at position $i$.
In our model, the calculation of transition score $\psi_{\bar{\vec{y}}_i, \bar{\vec{y}}_{i+1}}$ is similar to the one in CRF\footnote{We calculate the transition parameters among five sub-tags $BIOES$ for targets.}.
{\color{black}For the factorized feature score $\Phi_{\vec{y}_i}(\vec{x},i)$, we will explain computation details based on a simple LSTM-based neural network in the following two subsections. }
Such a factorized feature score is able to encode both token-level features as in standard CRF, segment-level features as in Semi-Markov CRF as well as the interaction among a target, its sentiment and an opinion span in a triplet.

	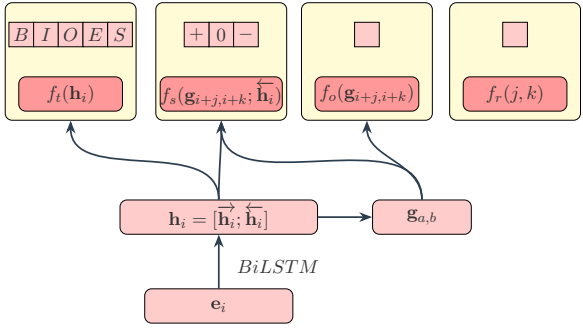
\begin{figure}[t!]
		\centering
		\adjustbox{max width=1.0\linewidth}{
			\begin{tikzpicture}[node distance=1.0mm and 1.0mm, >=Stealth, 
			wordnode/.style={draw=none, minimum height=5mm, inner sep=0pt},
			embnode/.style={draw=black, rounded corners, fill=red!20, minimum height=7mm, minimum width=30mm, inner sep=0pt},
			layernode/.style={draw=black, rounded corners, fill=red!20, minimum height=7mm, minimum width=20mm, inner sep=0pt},
			scoresnode/.style={draw=black, rounded corners, fill=red!40, minimum height=7mm, minimum width=21mm, inner sep=0pt},
			scorenode/.style={draw=black, fill=red!20, minimum height=5mm, minimum width=5mm, inner sep=0pt},
			chainLine/.style={line width=1pt,-, color=fontgray},
			scopebox/.style={draw=black, rounded corners, fill=yellow!20, minimum height = 24mm},
			scorebox/.style={draw=black,  rounded corners, fill=yellow!20, minimum height = 10mm, minimum width = 10mm},
			btag/.style={shape=circle, draw=black, rounded corners, fill=orange!20, minimum height=8mm, inner sep=0pt},
			atag/.style={shape=circle, draw=black, rounded corners, fill=orange!20, minimum height=8mm, inner sep=0pt},
			etag/.style={shape=circle, draw=black, rounded corners, fill=orange!20, minimum height=8mm, inner sep=0pt}
			]

			\node [embnode] (emb)  [] {\color{black!80}\textsc{$\vec{e}_i$}};

			\node [wordnode, above=of emb, xshift=12mm, yshift=1mm] (bilstm)  [] {\color{black!80}\textsc{$BiLSTM$}};
			
			\node [layernode, above=of emb, minimum width=40mm, yshift=10mm, xshift=0mm] (lstm)  [] {\color{black!80}\textsc{$\vec{h}_i = [\overrightarrow{\vec{h}_i};\overleftarrow{\vec{h}_i}]$}}; 
			\node [layernode, right=of lstm, xshift=10mm] (segmental)  [] {\color{black!80}\textsc{$\vec{g}_{a,b} $}};

			\node [scoresnode, above=of lstm, yshift=17mm, xshift=-30mm] (f_t)  [] {\color{black!80}\textsc{$f_t(\vec{h}_i)$}};
			\node [scoresnode, above=of lstm, yshift=17mm, xshift=0.5mm] (f_s)  [] {\color{black!80}\textsc{$f_s(\vec{g}_{i + j, i+ k};\overleftarrow{\vec{h}_i})$}};
			\node [scoresnode, above=of lstm, yshift=17mm, xshift=30mm] (f_g)  [] {\color{black!80}\textsc{$f_o(\vec{g}_{i+j, i+k})$}};
			\node [scoresnode, above=of lstm, yshift=17mm, xshift=60mm] (f_r)  [] {\color{black!80}\textsc{$f_r(j,k)$}};

			\node [scorenode, above=of f_s, yshift=5mm, fill=red!20, xshift=0mm] (f_s_plus)  [] {\color{black!80}\textsc{$0$}};
			\node [scorenode, above=of f_s, yshift=5mm, fill=red!20, xshift=-5mm] (f_s_0)  [] {\color{black!80}\textsc{$+$}};
			\node [scorenode, above=of f_s, yshift=5mm, fill=red!20, xshift=5mm] (f_s_minus)  [] {\color{black!80}\textsc{$-$}};

			\node [scorenode, above=of f_t, yshift=5mm, fill=red!20, xshift=-10mm] (f_t_B)  [] {\color{black!80}\textsc{$B$}};
			\node [scorenode, above=of f_t, yshift=5mm, fill=red!20, xshift=-5mm] (f_t_M)  [] {\color{black!80}\textsc{$I$}};
			\node [scorenode, above=of f_t, yshift=5mm, fill=red!20, xshift=0mm] (f_t_E)  [] {\color{black!80}\textsc{$O$}};
			\node [scorenode, above=of f_t, yshift=5mm, fill=red!20, xshift=5mm] (f_t_S)  [] {\color{black!80}\textsc{$E$}};
			\node [scorenode, above=of f_t, yshift=5mm, fill=red!20, xshift=10mm] (f_t_O)  [] {\color{black!80}\textsc{$S$}};

			\node [scorenode, above=of f_g, yshift=5mm, fill=red!20, xshift=0mm] (f_g_x)  [] {\color{black!80}\textsc{}};
			
			\node [scorenode, above=of f_r, yshift=5mm, fill=red!20, xshift=0mm] (f_r_x)  [] {\color{black!80}\textsc{}};


			\begin{pgfonlayer}{background}
			\node [scopebox, above=of f_t, xshift=0mm, yshift=-10mm, text height=5mm, minimum width=26.5mm] (s_t)  [] {\color{blue!80}\textbf{\textsc{}}};
			\node [scopebox, above=of f_s, xshift=0mm, yshift=-10mm, text height=5mm, minimum width=26.5mm] (s_s)  [] {\color{blue!80}\textbf{\textsc{}}};
			\node [scopebox, above=of f_g, xshift=0mm, yshift=-10mm, text height=5mm, minimum width=26.5mm] (s_g)  [] {\color{blue!80}\textbf{\textsc{}}};
			\node [scopebox, above=of f_r, xshift=0mm, yshift=-10mm, text height=5mm, minimum width=26.5mm] (s_r)  [] {\color{blue!80}\textbf{\textsc{}}};
						
			\end{pgfonlayer}

			\draw[chainLine,->]   (emb) to[out=90,in=-90] (lstm);
			\draw[chainLine,->]   (lstm) to[out=0,in=180] (segmental);

			\draw[chainLine,->]   (lstm) to[out=90,in=-90] (s_t);
			\draw[chainLine,->]   (lstm) to[out=90,in=-90] (s_s);
			\draw[chainLine,->]   (segmental) to[out=90,in=-90] (s_s);
			\draw[chainLine,->]   (segmental) to[out=90,in=-90] (s_g);
			
			\end{tikzpicture} 
		}
		\caption{Neural Module for Feature Score}
		\label{fig:neural}
	\end{figure}
\subsubsection{Neural Module}

We deploy a simple LSTM-based neural architecture for learning features.
Given an input token sequence $\vec{x}=\{x_1,x_2,\cdots,x_{n}\}$ of length $n$, we first obtain the embedding sequence $\{\vec{e}_1,\vec{e}_2, \cdots, \vec{e}_n\}$. 
As illustrated in Figure~\ref{fig:neural}, we then apply a bi-directional LSTM on the embedding sequence and obtain the hidden state $\vec{h}_i$ for each position $i$, which could be represented as:
\setlength{\abovedisplayskip}{4pt} \setlength{\abovedisplayshortskip}{4pt}
\setlength{\belowdisplayskip}{4pt} \setlength{\belowdisplayshortskip}{4pt}
\begin{equation}
    \vec{h}_i=[\overrightarrow{\vec{h}_i};\overleftarrow{\vec{h}_i}]
\end{equation}
where $\overrightarrow{\vec{h}_i}$ and $\overleftarrow{\vec{h}_i}$ are the hidden states of the forward and backward LSTMs respectively.


Motivated by~\cite{wang2016graph, stern2017minimal}, we calculate the segment representation $\vec{g}_{a,b}$ for an opinion span with {\color{black}boundaries of $a$ and $b$ (both inclusive)} as follows:
\setlength{\abovedisplayskip}{4pt} \setlength{\abovedisplayshortskip}{4pt}
\setlength{\belowdisplayskip}{4pt} \setlength{\belowdisplayshortskip}{4pt}
\begin{equation}
    \vec{g}_{a,b} = [\overrightarrow{\vec{h}}_{b} - \overrightarrow{\vec{h}}_{a-1}; \overleftarrow{\vec{h}}_{a} - \overleftarrow{\vec{h}}_{b+1}]
\end{equation}
where $\overrightarrow{\vec{h}}_{0} = \vec{0} $, $\overleftarrow{\vec{h}}_{n+1}=\vec{0}$ and $1\leq a \leq b \leq n. $

\subsubsection{Factorized Feature Score}
We explain how to compute the factorized feature scores (the second part of Equation~\ref{eq:score}) for the position-aware  tagging scheme based on the neural architecture described above.
Such factorized feature scores involve 4 types of scores, as illustrated in the solid boxes appearing in Figure~\ref{fig:neural} (top).

Basically, we calculate the factorized feature score for the tag $\vec{y}_i$ as follows:
\setlength{\abovedisplayskip}{4pt} \setlength{\abovedisplayshortskip}{4pt}
\setlength{\belowdisplayskip}{4pt} \setlength{\belowdisplayshortskip}{4pt}
\begin{equation}
    \Phi_{\vec{y}_{i}}(\vec{x},i) = f_t(\vec{h}_i)_{\bar{\vec{y}}_i}
\end{equation}
where the linear layer $f_t$ is used to calculate the score for local context for targets.
Such a linear layer takes the hidden state $\vec{h}_i$ as the input and returns a vector of length $5$, with each value in the vector indicating the score of the corresponding sub-tag among $BIOES$. 
The subscript $\bar{\vec{y}}_i$ indicates the index of such a sub-tag.

When $\vec{y}_i \in \{B^{\epsilon}_{j,k}, S^{\epsilon}_{j,k}\}$, we need to calculate 3 additional factorized feature scores for capturing structural information by adding them to the basic score as follows:
\setlength{\abovedisplayskip}{3pt} \setlength{\abovedisplayshortskip}{3pt}
\setlength{\belowdisplayskip}{3pt} \setlength{\belowdisplayshortskip}{3pt}
\begin{align}
    \label{eq:triplet}
    &\Phi_{\vec{y}_{i}}(\vec{x},i)  \pluseq \\
    & f_s([\vec{g}_{i+j,i+k};\overleftarrow{\vec{h}_{i}}])_{\epsilon} + f_o(\vec{g}_{i+j,i+k})  + f_r(j,k) \nonumber
\end{align}
Note that the subscript of the variable $\vec{g}$ is represented as $i+j,i+k$ which are the absolute positions since $j,k$ are the offsets. We explain such 3 additional factorized scores appearing in Equation~\ref{eq:triplet}. 


\begin{table*}[t]
\centering
\scalebox{0.64}{
\begin{tabular}{lrrrr|rrrr|rrrr|rrrr}
\toprule 
\multirow{2}{*}{\textbf{Dataset}} & \multicolumn{4}{c}{\texttt{14Rest}} & \multicolumn{4}{c}{\texttt{14Lap}} & \multicolumn{4}{c}{\texttt{15Rest}} & \multicolumn{4}{c}{\texttt{16Rest}} \\  
& \#S & \# + & \# 0 & \# - &  \#S & \# + & \# 0 & \# - & \#S & \# + & \# 0 & \# - & \#S & \# + & \# 0 & \# -  \\ \midrule
\textbf{Train} & 1266 & 1692 & 	166 & 	480 & 906  & 817 & 	126 & 	517 & 605 & 783 & 	25 & 	205  & 857 &  1015 & 	50 & 	329 \\ 
\textbf{Dev} & {\color{white}0,}310 & 404 & 	54 & 	119  & 219 &169 & 	36 & 	141 & 148 &185 & 	11 & 	53 & 210 &252 & 	11 & 	76 \\
\textbf{Test} & {\color{white}0,} 492 & 773 & 	66 & 	155 & 328 &364 & 	63 & 	116    & 322 &317 & 	25 & 	143& 326& 407 & 	29 & 	78  \\
\bottomrule
\end{tabular}
}
\vspace{-1mm}
\caption{Statistics of 4 datasets. (\#S denotes number of sentences, and \# +, \# 0, \# - denote numbers of positive, neutral and negative triplets respectively.)}
\label{tab:dataset}
\vspace{-1mm}
\end{table*}

\squishlist
\setlength\itemsep{0.5mm}

    \item $f_s([\vec{g}_{i+j,i+k};\overleftarrow{\vec{h}_{i}}])_{\epsilon}$ calculates the score for the sentiment.
     A linear layer $f_s$ takes the concatenation of the segment representation $\vec{g}_{i+j, i+k}$ for an opinion span and the local context $\overleftarrow{\vec{h}_{i}}$ for a target, since we believe that the sentiment is mainly determined by the opinion span as well as the target phrase itself.
    Note that we only use the backward hidden state $\overleftarrow{\vec{h}_{i}}$ here, because the end position of a target is not available in the tag and the target phrase appears on the right of this position $i$.
    The linear layer $f_s$ returns a vector of length $3$, with each value representing the score of a certain polarity of $+,0,-$. 
    The subscript $\epsilon$ indicates the index of such a polarity.
    

    \item $f_o(\vec{g}_{i+j,i+k})$ is used to calculate a score for an opinion span.
     A linear layer $f_o$ takes the segment representation $\vec{g}_{i+j, i+k}$ of an opinion span and returns one number representing the score of an opinion span. 

    \item $f_r(j,k)$ is used to calculate a score for offsets, since we believe the offset is an important feature.
     A linear layer $f_r$ returns one number representing the score of offsets $j,k$ which again are the distances between a target and two ends of the opinion span. 
    Here, we introduce the offset embedding $\vec{w_r}$ randomly initialized for encoding different offsets.
    Specifically, we calculate the score as follows\footnote{We use $\min{(j,k)}$ since we care the offset between the starting positions of an opinion span and a target.}:
    \setlength{\abovedisplayskip}{4pt} \setlength{\abovedisplayshortskip}{4pt}
\setlength{\belowdisplayskip}{4pt} \setlength{\belowdisplayshortskip}{4pt}
    \begin{equation}
        f_r(j,k) = W_{r} \vec{w_r}[\min{(j,k)}] + b_{r}
    \end{equation}
    where $W_r$ and $b_r$ are learnable parameters.

\squishend
\subsection{One Target for Multiple Opinion Spans}
\label{sec:variant}

The approach \textbf{JET} described above allows multiple targets to point to the same opinion span.
One potential issue is that such an approach is not able to handle the case where one target is associated with multiple opinion spans. 
To remedy such an issue, we could swap a target and an opinion span to arrive at a new model as a model variant, since they are both text spans which are characterized by their boundaries.
Specifically, in such a model variant, we still use the extended tags $B^{\epsilon}_{j,k}$ and $S^{\epsilon}_{j,k}$, where we use sub-tags $BIOES$ to encode an opinion span, the offsets $j,k$ for the target and $\epsilon$ for the sentiment polarity.
We use a similar procedure for the feature score calculation.

\begin{figure}[t!]
		\centering
				\adjustbox{max width=1.0\linewidth}{
			\begin{tikzpicture}[node distance=1.0mm and 1.0mm, >=Stealth, 
			wordnode/.style={draw=none, minimum height=5mm, inner sep=0pt},
			chainLine/.style={line width=1pt,->, color=blue},
			opinionbox/.style={draw=black, rounded corners, fill=yellow!20, dashed},
			targetbox/.style={draw=black, rounded corners, fill=red!20},
			BStag/.style={shape=circle, draw=black, rounded corners, fill=orange!20, minimum height=12mm, inner sep=0pt},
			MEOtag/.style={shape=circle, draw=black, rounded corners, fill=orange!20, minimum height=6mm, inner sep=0pt}
			]

			\matrix (sent1) [matrix of nodes, nodes in empty cells, execute at empty cell=\node{\strut};]
			{
				\textbf{food} & [1mm] was &[3mm] so & [4mm]  so & [3mm] but &  [2mm]  excited & [2mm] to & [1mm] see & [1mm] many & [1mm] \textbf{vegan} & [1mm] \textbf{options}\\
			};
			
			
			\foreach \pos in {6}
				\node [BStag, below=of sent1-1-\pos, yshift=-2.4mm] (tag\pos)  [] {\color{black!80}\textsc{$S^{\scriptscriptstyle +}_{\scriptscriptstyle 4,5}$}};

			\foreach \pos/\yoffset in {{2/-5.45},{5/-5.45},{1/-5.45},{7/-5.45},{8/-5.45},{9/-4.5},{10/-4.5},{11/-4.5}}
				\node [MEOtag, below=of sent1-1-\pos, yshift=\yoffset mm] (tag\pos)  [] {\color{black!80}\textsc{$O$}};
			
			\node [BStag, below=of sent1-1-3, yshift=-2.4mm] (tag3)  [] {\color{black!80}\textsc{$B^{\scriptscriptstyle 0}_{\scriptscriptstyle -2,-2}$}};
			
			\node [MEOtag, below=of sent1-1-4, yshift=-5.45mm] (tag4)  [] {\color{black!80}\textsc{$E$}};

			\foreach \pos/\npos in {{1/2},{2/3},{3/4},{4/5},{5/6},{6/7},{7/8},{8/9},{9/10},{10/11}}
				\draw[-]  (tag\pos) to[out=0,in=180] (tag\npos);

			\begin{pgfonlayer}{background}
			
			\node [targetbox, above=of sent1-1-1, yshift=-7mm, text height=-2mm, minimum height = 10mm, minimum width=10mm] (e1)  [] {\color{blue!80}\textbf{\textsc{0}}};	
			\node [opinionbox, above=of sent1-1-3, xshift=5mm, yshift=-6mm, text height=-5mm, minimum height = 10mm, minimum width=20mm] (o1)  [] {\color{blue!80}\textbf{\textsc{}}};
			
			\draw[chainLine,->]   (e1) to[out=45,in=135] (o1);

			\node [targetbox, above=of sent1-1-10, xshift=8mm, yshift=-7mm, text height=-2mm, minimum height = 10mm, minimum width=30mm] (e2)  [] {\color{blue!80}\textbf{\textsc{+}}};			
			\node [opinionbox, above=of sent1-1-6, xshift=0mm, yshift=-7mm, text height=-5mm, minimum height = 10mm, minimum width=15mm] (o2)  [] {\color{blue!80}\textbf{\textsc{}}};		
			
			\draw[chainLine,->]  (e2) to[out=160,in=30] (o2);
				
			\end{pgfonlayer}

			\end{tikzpicture} 
		}		
		\caption{The gold tagging sequence of {\bf JET}\textsuperscript{o} for the example sentence.}
		\vspace{-1mm}
		\label{fig:tagging_o}
	\end{figure}
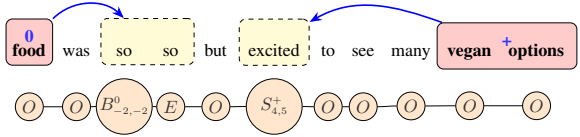



To differentiate with our first model, we name our first model as \textbf{JET}$^t$ and such a model variant as \textbf{JET}$^o$.
The superscripts $t$ and $o$ indicate the use of the sub-tags $B$ and $S$ to encode a target and an opinion span respectively.
Figure~\ref{fig:tagging_o} presents the gold tagging sequence of \textbf{JET}$^o$.

\subsection{Training and Inference}
The loss function $\mathcal{L}$ for the training data $D$ is defined as:
\setlength{\abovedisplayskip}{-2pt} \setlength{\abovedisplayshortskip}{-2pt}
\setlength{\belowdisplayskip}{3pt} \setlength{\belowdisplayshortskip}{3pt}
\begin{equation}
    \mathcal{L} = -\sum_{(\vec{x},\vec{y}) \in D} \log{p(\vec{y}|\vec{x})}.
\end{equation}
The overall model is analogous to that of a neural CRF~\cite{NIPS2009_3869,do2010neural,lample2016neural}; hence the inference and decoding follow standard marginal and MAP inference\footnote{\textcolor{black}{See the Appendix for detailed algorithm.}} procedures.
For example, the prediction of $\vec{y}$ follows the Viterbi-like MAP inference procedure during decoding.
Notice that the number of labels at each position under the  position-aware tagging scheme is $O(M^2)$, since we need to compute segment representation for text spans of lengths within $M$.
Hence, the time complexity for inference is $O(nM^2)$.
{\color{black}When $M \ll n$ (empirically, we found $n$ can be up to 80 in our datasets, and we set $M \in [2,6]$), this complexity is better than the existing work with complexity $O(n^2)$~\cite{peng2019knowing}.}




\section{Experiments}

\subsection{Data}
\textcolor{black}{ 
We refine the dataset previously created by~\citet{peng2019knowing}\footnote{ \url{https://github.com/xuuuluuu/SemEval-Triplet-data}}.
We call our refined dataset  ASTE-Data-V2, and the original version as ASTE-Data-V1\footnote{We also report the results on ASTE-Data-V1 in the Appendix.}.
Note that ASTE-Data-V1 does not contain  cases where one opinion span is associated with multiple targets.
For example, there are two targets, ``\textit{service}'' and ``\textit{atmosphere}'', in the sentence ``\textit{Best service and atmosphere}''. The opinion span ``\textit{Best}'' is associated with such two targets, resulting in two triplets.
However, we found that not all such triplets are explicitly annotated in ASTE-Data-V1.
We refine the dataset with these additional missing triplets in our dataset ASTE-Data-V2\footnote{We also remove triplets with sentiment originally labeled as ``conflict'' by SemEval.}.
}




Table~\ref{tab:dataset} presents the detailed statistics for 4 datasets.\footnote{See the Appendix for more statistics.}
{\texttt{14Rest}}, {\texttt{15Rest}}, {\texttt{16Rest}} are the datasets of restaurant  domain and  {\texttt{14Lap}} is of laptop  domain.
Such datasets were all created based on the datasets originally released by SemEval~\cite{pontiki-EtAl:2014:SemEval,pontiki-etal-2015-semeval,pontiki2016semeval, fan-etal-2019-target}.

\begin{table*}[!t]

    \centering
    \scalebox{0.62}{
    \begin{tabular}{lcccc|cccc|cccc|cccc}
    \toprule
      \multirow{2}{*}{\textbf{Models}} & \multicolumn{4}{c}{\texttt{14Rest}} & \multicolumn{4}{c}{\texttt{14Lap}} & \multicolumn{4}{c}{\texttt{15Rest}}  & \multicolumn{4}{c}{\texttt{16Rest}}    \\ 
      &Dev $F_1$ & $P.$ & $R.$ & $F_1$&Dev $F_1$ & $P.$ & $R.$ & $F_1$&Dev $F_1$ & $P.$ & $R.$ & $F_1$ &Dev $F_1$ & $P.$ & $R.$ & $F_1$ \\ \midrule
       
        \textbf{CMLA+} & -& 39.18 & 47.13 & 42.79 &-&  30.09 & 36.92 & 33.16 &-&  34.56 & 39.84 & 37.01 &-&  41.34 & 42.10 & 41.72          \\
        \textbf{RINANTE+} &-&  31.42 & 39.38 & 34.95 &-&  21.71 & 18.66 & 20.07  &-&  29.88 & 30.06 & 29.97 &-&  25.68 & 22.30 & 23.87  \\
        \textbf{Li-unified-R} &-&  41.04 & 67.35 & 51.00 &-&  40.56 & 44.28 & 42.34 &-&  44.72 & 51.39 & 47.82 &-&  37.33 & 54.51 & 44.31 \\
        \citet{peng2019knowing} &-&  43.24 & 63.66 & 51.46 &-&  37.38 & 50.38 & 42.87 &-&  48.07 & 57.51 & 52.32 &-&  46.96 & 64.24 & 54.21 \\ 
        \midrule
        \textbf{JET}$^t$ $(M=2)$& 
        45.67 & 72.46 & 32.29 & 44.68 & 35.69 & 57.39 & 24.31 & 34.15 & 42.34 &  64.81 & 28.87 & 39.94 & 43.27 & 68.75 & 38.52 & 49.38 \\
        \textbf{JET}$^t$ $(M=3)$&
        50.87 & 70.02 & 42.76 & 53.09 & 42.34 &  56.86 & 31.31 & 40.38 & 52.02 & 59.87 & 36.91 & 45.66 & 52.13 & 67.22 & 47.47 & 55.64 \\
        \textbf{JET}$^t$ $(M=4)$ &
        50.31 & 69.67 & 47.38 & 56.41 & 45.90 & 48.77 & 32.78 & 39.21 & 52.50 & 64.50 & 40.82 & 50.00 & 57.69 & 64.64 & 47.67 & 54.87 \\
        
        \textbf{JET}$^t$ $(M=5)$&52.41 & 62.23 & 48.39 & 54.44 & \underline{48.26} &  54.84 & 34.44 & \bf{42.31} & 54.97 & 55.67 & 43.51 & 48.84 & 57.83 & 61.63 & 48.44 & 54.25 \\
        \textbf{JET}$^t$ $(M=6)$&\underline{53.14}& 66.76&49.09& \bf{56.58}& 47.68& 52.00& 35.91 &42.48 &\underline{55.06}& 59.77&42.27& \bf{49.52}& \underline{58.45} &63.59&50.97&\bf{56.59}\\[1.5mm]
        \textbf{JET}$^o$ $(M=2)$&
        41.72 & 66.89 & 30.48 & 41.88 & 36.12 & 54.34 & 21.92 & 31.23 & 43.39 & 52.31 & 28.04 & 36.51 & 43.24 & 63.86 & 35.41 & 45.56 \\
        
        \textbf{JET}$^o$ $(M=3)$ &
        49.41 & 65.29 & 41.45 & 50.71 & 41.95 & 58.89 & 31.12 & 40.72 & 48.72 & 58.28 & 34.85 & 43.61 & 53.36 & 72.40 & 47.47 & 57.34 \\
        
        \textbf{JET}$^o$ $(M=4)$&
        51.56 & 67.63 & 46.88 & 55.38 & 45.66 & 54.55 & 35.36 & 42.91 & 56.73 & 58.54 & 43.09 & 49.64 & 58.26 & 69.81 & 49.03 & 57.60 \\
        
        \textbf{JET}$^o$ $(M=5)$ &
        53.35 & 71.49 & 47.18 & 56.85 & \underline{45.83} & 55.98 & 35.36 & \bf{43.34} & 59.57 & 61.39 & 40.00 & 48.44 & 55.92 & 66.06 & 49.61 & 56.67 \\
        
        \textbf{JET}$^o$ $(M=6)$&\underline{53.54}& 61.50 &55.13&\bf{58.14}& 45.61 &53.03	&33.89&	41.35 &\underline{60.97}&64.37&	44.33&	\bf{52.50}& \underline{60.90} & 70.94 & 57.00 & \bf{63.21}\\ [1.5mm] 
        \multicolumn{4}{l}{\textbf{+ Contextualized Word Representation (BERT)}}& & &&&&&&&&&&&\\
        \textbf{JET}$^t$ $(M=6)_{\scriptscriptstyle \text{+ BERT}}$&
        56.00 & 63.44 & 54.12 & 58.41 & 50.40 & 53.53 & 43.28 & 47.86 & 59.86 & 68.20 & 42.89 & 52.66 & 60.67 & 65.28 & 51.95 & 57.85 \\
        \textbf{JET}$^o$ $(M=6)_{\scriptscriptstyle \text{+ BERT}}$ &
        56.89 & 70.56 & 55.94 & 62.40 & 48.84 & 55.39 & 47.33 & 51.04 & 64.78 & 64.45 & 51.96 & 57.53 & 63.75 & 70.42 & 58.37 & 63.83 \\
        
  \bottomrule
    \end{tabular}
    }
    \vspace{-1.5mm}
    \caption{Main results on our refined dataset ASTE-Data-V2. The underlined scores indicate the best results on the dev set, and the highlighted scores are the corresponding test results. \textcolor{black}{The experimental results on the previous released dataset ASTE-Data-V1 can be found in the Appendix.}}
    \label{tab:main_results}
    \vspace{-4mm}
\end{table*}

\subsection{Baselines}
Our \textbf{JET} approaches are compared with the following baselines using pipeline. 

\squishlist
    \item \textbf{RINANTE+}~\cite{peng2019knowing} modifies \textbf{RINANTE}~\cite{dai2019neural} which is designed based on {LSTM-CRF}~\cite{lample2016neural}, to co-extract targets with sentiment, and opinion spans.
    Such an approach also fuses mined rules as weak supervision to capture dependency relations of words in a sentence at the first stage. 
    At the second stage, it generates all the possible triplets and applies a classifier based on MLP on such triplets to determine if each triplet is valid or not.
    
    \item \textbf{CMLA+}~\cite{peng2019knowing} modifies \textbf{CMLA}~\cite{wang2017coupled} which leverages attention mechanism to capture dependencies among words, to co-extract targets with sentiment, and opinion spans at the first stage. 
    At the second stage, it uses the same method to obtain all the valid triplets as \textbf{RINANTE+}.
    \item \textbf{Li-unified-R}~\cite{peng2019knowing} modifies the model~\cite{li2019unified} to extract targets with sentiment, as well as opinion spans respectively based on a customized multi-layer LSTM neural architecture.
    At the second stage, it uses the same method to obtain all the valid triplets as \textbf{RINANTE+}.
    
    \item \citet{peng2019knowing} proposed an approach motivated by \textbf{Li-unified-R} to co-extract targets with sentiment, and opinion spans simultaneously.
    Such an approach also fuses GCN to capture dependency information to facilitate the co-extraction. 
    At the second stage, it uses the same method to obtain all the valid triplets as \textbf{RINANTE+}.
    

\squishend


\subsection{Experimental Setup}
Following the previous work~\cite{peng2019knowing}, we use pre-trained 300d  GloVe~\cite{pennington2014glove} to initialize the word embeddings. 
We use 100 as the embedding size of $\vec{w_r}$ (offset embedding).
We use the bi-directional LSTM with the hidden size $300$.
For experiments with contextualised representation, we adopt the pre-trained language model BERT \cite{devlin2019bert}.  Specifically, we use bert-as-service \cite{xiao2018bertservice} to generate the contextualized word embedding without fine-tuning. We use the representation from the last layer of the uncased version of BERT base model for our experiments.

Before training, we discard any instance from the training data that contains triplets with offset larger than $M$.
We train our model for a maximal of 20 epochs using Adam~\cite{kingma2014adam} as the optimizer with batch size $1$ and dropout rate $0.5$\footnote{See the Appendix for experimental details. We use a different dropout rate $0.7$ on the dataset {\texttt{14Lap}} based on preliminary results since the domain is different from the other 3 datasets.}.
We select the best model parameters based on the best $F_1$ score on the development data and apply it to the test data for evaluation.

Following the previous works, we report the \textit{precision} ($P.$), \textit{recall} ($R.$) and \textit{$F_1$} scores for the correct triplets. 
Note that a correct triplet requires the boundary\footnote{We define a boundary as the beginning and ending positions of a text span.} of the target, the boundary of the opinion span, and the target sentiment polarity to be all correct at the same time.



\subsection{Main Results}
Table~\ref{tab:main_results} presents the main results, where all the baselines as well as our models with different maximum offsets $M$ are listed.
In general, our joint models \textbf{JET}$^t$ and \textbf{JET}$^o$, which are selected based on the best $F_1$ score on the dev set, are able to outperform the most competitive baseline of~\citet{peng2019knowing} on the 4 datasets {\texttt{14Rest}}, {\texttt{15Rest}}, {\texttt{16Rest}}, and {\texttt{14Lap}}.
\textcolor{black}{Specifically, the best models selected from \textbf{JET}$^t$ and \textbf{JET}$^o$ outperform~\citet{peng2019knowing} significantly\footnote{We have conducted significance test using the bootstrap resampling method~\cite{koehn-2004-statistical}.} on {\texttt{14Rest}} and {\texttt{16Rest}} datasets with $p<10^{-5}$ respectively.}
Such results imply that our joint models \textbf{JET}$^t$ and \textbf{JET}$^o$ are more capable of capturing interactions among the elements in triplets than those pipeline approaches.
In addition, we observe a general trend from the results that the $F_1$ score increases as $M$ increases on the 4 datasets when $M \leq 5$. \textcolor{black}{We observe that the performance of \textbf{JET}$^t$ and \textbf{JET}$^o$ on the dev set of {\texttt{14Lap}} drops when $M =6$.} 

For the dataset {\texttt{14Rest}}, \textbf{JET}$^o(M=6)$ achieves the best results on $F_1$ scores among all the \textbf{JET}$^o$ models.
Such a \textbf{JET}$^o(M=6)$ model outperforms the strongest baseline~\citet{peng2019knowing} by nearly $7$ $F_1$ points.
\textbf{JET}$^t(M=6)$ also achieves a good performance with 56.58 in terms of $F_1$ score.
Comparing results of our models to baselines, the reason why ours have better $F_1$ scores is that our models \textbf{JET}$^t(M \geq 4)$ and \textbf{JET}$^o(M \geq 4)$ both achieve improvements of more than $15$ precision points, while we maintain acceptable recall scores.
\textcolor{black}{Similar patterns of results on the datasets {\texttt{14Lap}}, {\texttt{15Rest}} and {\texttt{16Rest}} are observed, except that \textbf{JET}$^t(M =5)$ and \textbf{JET}$^o(M =5)$ achieves the best $F_1$ score on the dev set of  {\texttt{14Lap}}.}
Furthermore, we discover that the performance of both \textbf{JET}$^o$ and \textbf{JET}$^t$ on {\texttt{14Rest}} and {\texttt{16Rest}} datasets is better than on{\texttt{14Lap}} and {\texttt{15Rest}} datasets. Such a behavior can be explained by the large distribution differences of  positive, neutral and negative sentiment between the train and test set of the {\texttt{14Rest}} and {\texttt{16Rest}} datasets, shown in Table \ref{tab:dataset}.

Furthermore, we also conduct additional experiments on our proposed model with the contextualized word representation BERT.  Both \textbf{JET}$^t$ $(M=6)_{\scriptscriptstyle \text{+ BERT}}$ and \textbf{JET}$^o$ $(M=6)_{\scriptscriptstyle \text{+ BERT}}$ achieve new state-of-the-art performance on the four datasets.
\section{Analysis}
\subsection{Robustness Analysis}

We analyze the model robustness by assessing the performance on targets, opinion spans and offsets of different lengths for two models \textbf{JET}$^t (M=6)_{\scriptscriptstyle \text{+ BERT}}$ and \textbf{JET}$^o(M=6)_{\scriptscriptstyle \text{+ BERT}}$ on the four datasets.
 Figure~\ref{fig:f1_diff_lengths} shows the results on the {\texttt{14Rest}} dataset\footnote{See the Appendix for the statistics of accumulative percentage of different lengths for targets, opinion spans and  offsets.}.
\textcolor{black}{As we can see, \textbf{JET}$^o(M=6)_{\scriptscriptstyle \text{+ BERT}}$ is able to better extract triplets with targets of lengths $\leq 3$ than \textbf{JET}$^t(M=6)_{\scriptscriptstyle \text{+ BERT}}$.
Furthermore, 
\textbf{JET}$^o(M=6)_{\scriptscriptstyle \text{+ BERT}}$ achieves a better $F_1$ score for triplets whose opinion spans are of length $1$ and $4$.
However,  \textbf{JET}$^o(M=6)_{\scriptscriptstyle \text{+ BERT}}$ performs comparably to  \textbf{JET}$^t(M=6)_{\scriptscriptstyle \text{+ BERT}}$ for triplets whose opinion spans are of length $2$ and $3$.
In addition, \textbf{JET}$^o(M=6)_{\scriptscriptstyle \text{+ BERT}}$ is able to  outperform \textbf{JET}$^t(M=6)_{\scriptscriptstyle \text{+ BERT}}$ with offset of length $4$ and above.} \textcolor{black}{We also observe that the performance drops when the lengths of targets, opinion spans and offsets are longer. This confirms that modeling the boundaries are harder when their lengths are longer.}
Similar patterns of results are observed on {\texttt{14Lap}}, {\texttt{15Rest}}, and {\texttt{16Rest}}\footnote{See the Appendix for results on the other 3 datasets.}.

\begin{figure}[t!]
\centering
\begin{subfigure}{0.45\linewidth}
   \centering
    \begin{adjustwidth}{-0.4cm}{0.0cm}
    \begin{tikzpicture}
    \pgfplotsset{width=5cm,compat=1.9}
    \begin{axis}
    [
                title={},
                legend style={font=\fontsize{5}{1}\selectfont},
                legend style={at={(0.5,0.75)},anchor=south,legend columns=2}, 
                xticklabel style = {font=\fontsize{6}{1}\selectfont},
                yticklabel style = {font=\fontsize{6}{1}\selectfont},
                xmin=1, xmax=6,
                ymin=0, ymax=115,
                xtick={1,2,3,4,5,6,7},
                xticklabels = {1,2,3,4,5,6,$>=7$},
                ytick={0.0,20,40,60, 80,100},
            ]
    \addplot [mark=square, color=red] plot coordinates {
    (1,62.73) (2, 51.36) (3, 40.45) (4, 23.08) (5, 0)(6, 66.67)(7, 0)};
    \addplot [mark=diamond*, color=blue] plot coordinates {
    (1, 60.72) (2, 37.74) (3, 13.79)  (4, 36.36) (5,0) (6, 0)(7, 0) };
    \addplot [mark=o, color=black] plot coordinates {
    (1, 72.87) (2, 76.01) (3, 64.11) ( 4, 47.92) (5, 44.93) (6, 21.78)(7, 0) };
    \legend{target \\opinion Span\\ offset\\}
    \end{axis}
    \end{tikzpicture}
    \end{adjustwidth}
    \centering
    \vspace{-2mm}
    \caption{\textbf{JET}$^t (M=6)_{\scriptscriptstyle \text{+ BERT}}$}
\end{subfigure}
\hfill
\begin{subfigure}{0.45\linewidth}
   \centering
   \begin{adjustwidth}{-0.4cm}{0.0cm}
        \begin{tikzpicture}
        \pgfplotsset{width=5cm,compat=1.9}
        \begin{axis}
        [
                    title={},
                    legend style={font=\fontsize{5}{1}\selectfont}, 
                    legend style={at={(0.5,0.75)},anchor=south,legend columns=2}, 
                    xticklabel style = {font=\fontsize{6}{1}\selectfont},
                    yticklabel style = {font=\fontsize{6}{1}\selectfont},
                    xmin=1, xmax=6,
                    ymin=0, ymax=115,
                    xtick={1,2,3,4,5,6},
                    xticklabels = {1,2,3,4,5,6},
                    ytick={0.0,20,40,60, 80,100},
                ]
    \addplot [mark=square, color=red] plot coordinates {
    (1,67.79) (2, 51.77) (3, 43.59) (4, 14.29) (5, 0)(6, 0)(7, 0)};
    \addplot [mark=diamond*, color=blue] plot coordinates {
    (1, 64.57) (2, 36.56) (3, 14.81)  (4, 72.73) (5,0) (6, 0)(7, 0) };
    \addplot [mark=o, color=black] plot coordinates {
    (1, 70.79) (2, 78.59) (3, 63.49) ( 4, 66.67) (5, 58.65) (6, 43.96)(7, 0) };
        \legend{target \\opinion span\\ offset\\}
        \end{axis}
        \end{tikzpicture}
    \end{adjustwidth}
    \centering
    \vspace{-1.5mm}
   \caption{\textbf{JET}$^o (M=6)_{\scriptscriptstyle \text{+ BERT}}$}
\end{subfigure}
\centering
\vspace{-2.5mm}
\caption{$F_1(\%)$ scores ($y$-axis) of different lengths ($x$-axis) for targets, opinion spans and offsets on the dataset {\texttt{14Rest}}.}
\label{fig:f1_diff_lengths}
\vspace{-3mm}
\end{figure}
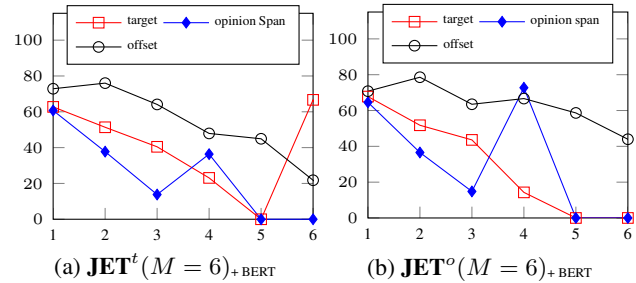

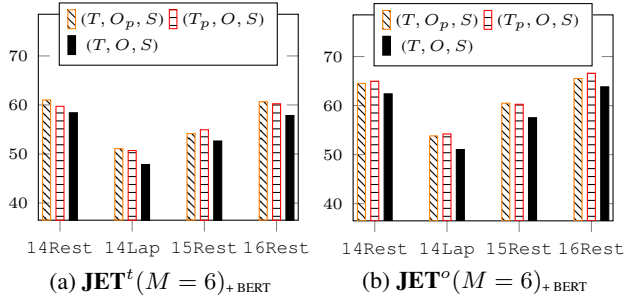
\begin{figure}[t!]
\centering
\begin{subfigure}{0.45\linewidth}
\centering
\begin{adjustwidth}{-0.4cm}{0cm}
\begin{tikzpicture}
\centering
\pgfplotsset{width=5cm,compat=1.9}
\begin{axis}[
    xtick={1,2,3,4},
    ymin=40, ymax=75,
    xticklabels = {{\texttt{14Rest}},{\texttt{14Lap}},{\texttt{15Rest}},{\texttt{16Rest}}},
    xticklabel style = {font=\fontsize{7}{1}\selectfont},
    yticklabel style = {font=\fontsize{6}{1}\selectfont},
    legend style={font=\fontsize{6}{1}\selectfont},
	enlargelimits=0.1,
	legend style={at={(0.5,0.75)},anchor=south,legend columns=2}, 
	ybar=2pt,
	bar width=3pt,
]

\addplot[orange,pattern=north west lines, area legend] coordinates {
(1, 61.02) (2, 51.12) (3, 54.18) ( 4, 60.67)
 };
 \addplot[red,pattern=horizontal lines, area legend] coordinates {
(1, 59.72) (2, 50.71) (3, 54.94) ( 4, 60.24)
 };
 \addplot[black,fill=black, area legend]  coordinates {
(1, 58.41) (2, 47.86) (3, 52.65) (4, 57.85) 
};

\legend{  {($T,O_p,S$)}, {($T_p,O,S$)}, {($T,O,S$)}}
\end{axis}
\end{tikzpicture}
\end{adjustwidth}
\vspace{-2mm}
\caption{\textbf{JET}$^t(M=6)_{\scriptscriptstyle \text{+ BERT}}$}
\label{fig:partial_t}
\end{subfigure}
\hfill
\begin{subfigure}{0.45\linewidth}
\centering
\begin{adjustwidth}{-0.4cm}{0cm}
\begin{tikzpicture}
\centering
\pgfplotsset{width=5cm,compat=1.9}
\begin{axis}[
    xtick={1,2,3,4},
    ymin=40, ymax=75,
    xticklabels = {{\texttt{14Rest}},{\texttt{14Lap}},{\texttt{15Rest}},{\texttt{16Rest}}},
    xticklabel style = {font=\fontsize{7}{1}\selectfont},
    yticklabel style = {font=\fontsize{6}{1}\selectfont},
    legend style={font=\fontsize{6}{1}\selectfont},
	enlargelimits=0.1,
	legend style={at={(0.5,0.75)},anchor=south,legend columns=2}, 
	ybar=2pt,
	bar width=3pt,
]

\addplot[orange,pattern=north west lines, area legend] coordinates {
(1, 64.53) (2, 53.82) (3, 60.50) ( 4, 65.53)
 };
 \addplot[red,pattern=horizontal lines, area legend] coordinates {
(1, 64.98) (2, 54.22) (3, 60.27) ( 4, 66.60)
 };
 \addplot[black,fill=black, area legend]  coordinates {
(1, 62.40) (2, 51.04) (3, 57.53) (4, 63.83) 
};
\legend{  {($T,O_p,S$)}, {($T_p,O,S$)}, {($T,O,S$)}}
\end{axis}
\end{tikzpicture}
\end{adjustwidth}
\vspace{-2mm}
\caption{\textbf{JET}$^o(M=6)_{\scriptscriptstyle \text{+ BERT}}$}
\label{fig:partial_o}
\end{subfigure}
\vspace{-2.5mm}
\caption{$F_1$ for different evaluation methods.}
\label{fig:partial}
\vspace{-4.5mm}
\end{figure}

\begin{table*}[h]
     \centering
     \scalebox{0.75}{
     \begin{tabular}{ c c c c }
     \toprule
      Gold & \citet{peng2019knowing} & \textbf{JET}$^t$ & \textbf{JET}$^o$ \\ 
    \midrule
            \scalebox{0.7}{
     		\begin{tikzpicture}[node distance=1.0mm and 1.0mm, >=Stealth, 
			wordnode/.style={draw=none, minimum height=2mm, inner sep=0pt},
			chainLine/.style={line width=1pt,->, color=blue, },
			opinionbox/.style={draw=black, rounded corners, fill=yellow!20, dashed},
			targetbox/.style={draw=black, rounded corners, fill=red!20},
			]

			\matrix (sent1) [matrix of nodes, ampersand replacement=\&, nodes in empty cells, execute at empty cell=\node{\strut};]
			{
				\textbf{Food} \& is \&  fresh \&  and \&  hot \&  ready \& to \& eat \\
			};

			\begin{pgfonlayer}{background}
			
			\node [targetbox, above=of sent1-1-1, yshift=-7mm, text height=-2mm, minimum height = 10mm, minimum width=10mm] (e1)  [] {\color{blue!80}\textbf{\textsc{+}}};	
			\node [opinionbox, above=of sent1-1-3, xshift=0mm, yshift=-6mm, text height=-5mm, minimum height = 7mm, minimum width=10mm] (o1)  [] {\color{blue!80}\textbf{\textsc{}}};
			
			\draw[chainLine,->]   (e1) to[out=30,in=140] (o1);

			\node [opinionbox, above=of sent1-1-5, xshift=0mm, yshift=-6mm, text height=-5mm, minimum height = 7mm, minimum width=8mm] (o2)  [] {\color{blue!80}\textbf{\textsc{}}};		
			
			\draw[chainLine,->]  (e1) to[out=32,in=140] (o2);
				
			\end{pgfonlayer}
			
			\end{tikzpicture} 
			}

      & 
         \scalebox{0.7}{
     		\begin{tikzpicture}[node distance=1.0mm and 1.0mm, >=Stealth, 
			wordnode/.style={draw=none, minimum height=2mm, inner sep=0pt},
			chainLine/.style={line width=1pt,->, color=blue},
			opinionbox/.style={draw=black, rounded corners, fill=yellow!20, dashed},
			targetbox/.style={draw=black, rounded corners, fill=red!20},
			]

			\matrix (sent1) [matrix of nodes, ampersand replacement=\&, nodes in empty cells, execute at empty cell=\node{\strut};]
			{
				\textbf{Food} \& is \&  fresh \&  and \&  hot \&  ready \& to \& eat \\
			};

			\begin{pgfonlayer}{background}
			
			\node [targetbox, above=of sent1-1-1, yshift=-7mm, text height=-2mm, minimum height = 10mm, minimum width=10mm] (e1)  [] {\color{blue!80}\textbf{\textsc{+}}};	
			\node [opinionbox, above=of sent1-1-3, xshift=0mm, yshift=-6mm, text height=-5mm, minimum height = 7mm, minimum width=10mm] (o1)  [] {\color{blue!80}\textbf{\textsc{}}};
			
			\draw[chainLine,->]   (e1) to[out=30,in=140] (o1);

			\node [opinionbox, above=of sent1-1-5, xshift=5mm, yshift=-6mm, text height=-5mm, minimum height = 7mm, minimum width=18mm] (o2)  [] {\color{blue!80}\textbf{\textsc{}}};		
			
			\draw[chainLine,->]  (e1) to[out=32,in=140] (o2);
				
			\end{pgfonlayer}
			
			\end{tikzpicture} 
			}
      & 
       \scalebox{0.7}{
     		\begin{tikzpicture}[node distance=1.0mm and 1.0mm, >=Stealth, 
			wordnode/.style={draw=none, minimum height=2mm, inner sep=0pt},
			chainLine/.style={line width=1pt,->, color=blue},
			opinionbox/.style={draw=black, rounded corners, fill=yellow!20, dashed},
			targetbox/.style={draw=black, rounded corners, fill=red!20},
			]

			\matrix (sent1) [matrix of nodes, ampersand replacement=\&, nodes in empty cells, execute at empty cell=\node{\strut};]
			{
				\textbf{Food} \& is \&  fresh \&  and \&  hot \&  ready \& to \& eat \\
			};

			\begin{pgfonlayer}{background}
			
			\node [targetbox, above=of sent1-1-1, yshift=-7mm, text height=-2mm, minimum height = 10mm, minimum width=10mm] (e1)  [] {\color{blue!80}\textbf{\textsc{+}}};	
			\node [opinionbox, above=of sent1-1-3, xshift=0mm, yshift=-6mm, text height=-5mm, minimum height = 7mm, minimum width=10mm] (o1)  [] {\color{blue!80}\textbf{\textsc{}}};
			
			\draw[chainLine,->]   (e1) to[out=30,in=140] (o1);

			
				
			\end{pgfonlayer}
			
			\end{tikzpicture} 
			}
     &
       \scalebox{0.7}{
     		\begin{tikzpicture}[node distance=1.0mm and 1.0mm, >=Stealth, 
			wordnode/.style={draw=none, minimum height=2mm, inner sep=0pt},
			chainLine/.style={line width=1pt,->, color=blue},
			opinionbox/.style={draw=black, rounded corners, fill=yellow!20, dashed},
			targetbox/.style={draw=black, rounded corners, fill=red!20},
			]

			\matrix (sent1) [matrix of nodes, ampersand replacement=\&, nodes in empty cells, execute at empty cell=\node{\strut};]
			{
				\textbf{Food} \& is \&  fresh \&  and \&  hot \&  ready \& to \& eat \\
			};

			\begin{pgfonlayer}{background}
			
			\node [targetbox, above=of sent1-1-1, yshift=-7mm, text height=-2mm, minimum height = 10mm, minimum width=10mm] (e1)  [] {\color{blue!80}\textbf{\textsc{+}}};	
			\node [opinionbox, above=of sent1-1-3, xshift=0mm, yshift=-6mm, text height=-5mm, minimum height = 7mm, minimum width=10mm] (o1)  [] {\color{blue!80}\textbf{\textsc{}}};
			
			\draw[chainLine,->]   (e1) to[out=30,in=140] (o1);

			\node [opinionbox, above=of sent1-1-5, xshift=0mm, yshift=-6mm, text height=-5mm, minimum height = 7mm, minimum width=8mm] (o2)  [] {\color{blue!80}\textbf{\textsc{}}};		
			
			\draw[chainLine,->]  (e1) to[out=32,in=140] (o2);
				
			\end{pgfonlayer}
			
			\end{tikzpicture} 
			}
      \\ 
       \midrule
        \scalebox{0.7}{
     	    	\begin{tikzpicture}[node distance=1.0mm and 1.0mm, >=Stealth, 
			wordnode/.style={draw=none, minimum height=2mm, inner sep=0pt},
			chainLine/.style={line width=1pt,->, color=blue},
			opinionbox/.style={draw=black, rounded corners, fill=yellow!20, dashed},
			targetbox/.style={draw=black, rounded corners, fill=red!20},
			]

			\matrix (sent1) [matrix of nodes, ampersand replacement=\&, nodes in empty cells, execute at empty cell=\node{\strut};]
			{
			   with \&  a \&  quaint \& [1mm] bar \&  and \& good \& [1mm] food  \\
			};

			\begin{pgfonlayer}{background}
			
			\node [targetbox, above=of sent1-1-4, yshift=-7mm, text height=-2mm, minimum height = 10mm, minimum width=10mm] (e1)  [] {\color{blue!80}\textbf{\textsc{0}}};	
			\node [opinionbox, above=of sent1-1-3, xshift=0mm, yshift=-6mm, text height=-5mm, minimum height = 7mm, minimum width=10mm] (o1)  [] {\color{blue!80}\textbf{\textsc{}}};
			
			\draw[chainLine,->]   (e1) to[out=150,in=45] (o1);

			\node [targetbox, above=of sent1-1-7, xshift=0mm, yshift=-7mm, text height=-2mm, minimum height = 10mm, minimum width=10mm] (e2)  [] {\color{blue!80}\textbf{\textsc{+}}};			
			\node [opinionbox, above=of sent1-1-6, xshift=0mm, yshift=-6mm, text height=-5mm, minimum height = 7mm, minimum width=9mm] (o2)  [] {\color{blue!80}\textbf{\textsc{}}};		
			
			\draw[chainLine,->]  (e2) to[out=150,in=45] (o2);
				
			\end{pgfonlayer}
			
			\end{tikzpicture} 
		}  
      &
              \scalebox{0.7}{
     	    	\begin{tikzpicture}[node distance=1.0mm and 1.0mm, >=Stealth, 
			wordnode/.style={draw=none, minimum height=2mm, inner sep=0pt},
			chainLine/.style={line width=1pt,->, color=blue},
			opinionbox/.style={draw=black, rounded corners, fill=yellow!20, dashed},
			targetbox/.style={draw=black, rounded corners, fill=red!20},
			]

			\matrix (sent1) [matrix of nodes, ampersand replacement=\&, nodes in empty cells, execute at empty cell=\node{\strut};]
			{
			   with \&  a \&  quaint \& [1mm] bar \&  and \& good \& [1mm] food  \\
			};

			\begin{pgfonlayer}{background}
			
			\node [targetbox, above=of sent1-1-4, yshift=-7mm, text height=-2mm, minimum height = 10mm, minimum width=10mm] (e1)  [] {\color{blue!80}\textbf{\textsc{0}}};	
			\node [opinionbox, above=of sent1-1-3, xshift=0mm, yshift=-6mm, text height=-5mm, minimum height = 7mm, minimum width=10mm] (o1)  [] {\color{blue!80}\textbf{\textsc{}}};
			
			\draw[chainLine,->]   (e1) to[out=150,in=45] (o1);
			\draw[chainLine,->]   (e1) to[out=30,in=135] (o2);

			\node [targetbox, above=of sent1-1-7, xshift=0mm, yshift=-7mm, text height=-2mm, minimum height = 10mm, minimum width=10mm] (e2)  [] {\color{blue!80}\textbf{\textsc{+}}};			
			\node [opinionbox, above=of sent1-1-6, xshift=0mm, yshift=-6mm, text height=-5mm, minimum height = 7mm, minimum width=9mm] (o2)  [] {\color{blue!80}\textbf{\textsc{}}};		
			
			\draw[chainLine,->]  (e2) to[out=150,in=45] (o2);
			\draw[chainLine,->]  (e2) to[out=140,in=55] (o1);
				
			\end{pgfonlayer}
			
			\end{tikzpicture} 
		}  
      &
              \scalebox{0.7}{
     	    	\begin{tikzpicture}[node distance=1.0mm and 1.0mm, >=Stealth, 
			wordnode/.style={draw=none, minimum height=2mm, inner sep=0pt},
			chainLine/.style={line width=1pt,->, color=blue},
			opinionbox/.style={draw=black, rounded corners, fill=yellow!20, dashed},
			targetbox/.style={draw=black, rounded corners, fill=red!20},
			]

			\matrix (sent1) [matrix of nodes, ampersand replacement=\&, nodes in empty cells, execute at empty cell=\node{\strut};]
			{
			   with \&  a \&  quaint \& [1mm] bar \&  and \& good \& [1mm] food  \\
			};

			\begin{pgfonlayer}{background}
			
			\node [targetbox, above=of sent1-1-4, yshift=-7mm, text height=-2mm, minimum height = 10mm, minimum width=10mm] (e1)  [] {\color{blue!80}\textbf{\textsc{0}}};	
			\node [opinionbox, above=of sent1-1-3, xshift=0mm, yshift=-6mm, text height=-5mm, minimum height = 7mm, minimum width=10mm] (o1)  [] {\color{blue!80}\textbf{\textsc{}}};
			
			\draw[chainLine,->]   (e1) to[out=150,in=45] (o1);

			\node [targetbox, above=of sent1-1-7, xshift=0mm, yshift=-7mm, text height=-2mm, minimum height = 10mm, minimum width=10mm] (e2)  [] {\color{blue!80}\textbf{\textsc{+}}};			
			\node [opinionbox, above=of sent1-1-6, xshift=0mm, yshift=-6mm, text height=-5mm, minimum height = 7mm, minimum width=9mm] (o2)  [] {\color{blue!80}\textbf{\textsc{}}};		
			
			\draw[chainLine,->]  (e2) to[out=150,in=45] (o2);
				
			\end{pgfonlayer}
			
			\end{tikzpicture} 
		}  
      &
        \scalebox{0.7}{
     	    	\begin{tikzpicture}[node distance=1.0mm and 1.0mm, >=Stealth, 
			wordnode/.style={draw=none, minimum height=2mm, inner sep=0pt},
			chainLine/.style={line width=1pt,->, color=blue},
			opinionbox/.style={draw=black, rounded corners, fill=yellow!20, dashed},
			targetbox/.style={draw=black, rounded corners, fill=red!20},
			]

			\matrix (sent1) [matrix of nodes, ampersand replacement=\&, nodes in empty cells, execute at empty cell=\node{\strut};]
			{
			   with \&  a \&  quaint \& [1mm] bar \&  and \& good \& [1mm] food  \\
			};

			\begin{pgfonlayer}{background}
			
			\node [targetbox, above=of sent1-1-4, yshift=-7mm, text height=-2mm, minimum height = 10mm, minimum width=10mm] (e1)  [] {\color{blue!80}\textbf{\textsc{0}}};	
			\node [opinionbox, above=of sent1-1-3, xshift=0mm, yshift=-6mm, text height=-5mm, minimum height = 7mm, minimum width=10mm] (o1)  [] {\color{blue!80}\textbf{\textsc{}}};
			
			\draw[chainLine,->]   (e1) to[out=150,in=45] (o1);

			\node [targetbox, above=of sent1-1-7, xshift=0mm, yshift=-7mm, text height=-2mm, minimum height = 10mm, minimum width=10mm] (e2)  [] {\color{blue!80}\textbf{\textsc{+}}};			
			\node [opinionbox, above=of sent1-1-6, xshift=0mm, yshift=-6mm, text height=-5mm, minimum height = 7mm, minimum width=9mm] (o2)  [] {\color{blue!80}\textbf{\textsc{}}};		
			
			\draw[chainLine,->]  (e2) to[out=150,in=45] (o2);
				
			\end{pgfonlayer}
			
			\end{tikzpicture} 
		}  
       \\

      \bottomrule
      \end{tabular}
      }
      \vspace{-2.5mm}
      \caption{Qualitative Analysis}
      \label{tag:qualitative}
    \vspace{-5mm}
\end{table*}

We also investigate the robustness on different evaluation methods, as presented in Figure~\ref{fig:partial}.
$T$ (Target), $O$ (Opinion Span) and $S$ (Sentiment) are the elements to be evaluated. 
The subscript $p$ on the right of an element in the legend denotes ``partially correct''.
We define two boundaries to be partially correct if such two boundaries overlap.
($T,O,S$) is the evaluation method used for our main results.
($T_p,O,S$) requires the boundary of targets to be partially correct, and the boundary of opinion spans as well as the sentiment to be exactly correct.
($T,O_p,S$) requires the boundary of opinion spans to be partially correct, and the boundary of targets as well as the sentiment to be exactly correct.
\textcolor{black}{The results based on ($T,O_p,S$) yield higher improvements in terms of $F_1$ points than results based on  ($T_p,O,S$), compared with ($T,O,S$) for \textbf{JET}$^t(M=6)_{\scriptscriptstyle \text{+ BERT}}$ except on \texttt{15Rest}.
The results based on ($T_p,O,S$) yield higher $F_1$ improvements than results based on  ($T,O_p,S$), compared with ($T,O,S$) for \textbf{JET}$^o(M=6)_{\scriptscriptstyle \text{+ BERT}}$ except on \texttt{15Rest}.
Such a comparison shows the boundaries of opinion spans or target spans may be better captured when the sub-tags $BIOES$ are used to model the opinion or target explicitly.}

\subsection{Qualitative Analysis}
To help us better understand the differences among these models, we present two example sentences selected from the test data as well as predictions by~\citet{peng2019knowing}, \textbf{JET}$^t$ and \textbf{JET}$^o$ in Table~\ref{tag:qualitative} \footnote{See the Appendix for more examples.}.
As we can see, there exist 2 triplets in the gold data in the first example.
\citet{peng2019knowing} predicts an incorrect opinion span \textit{``hot ready''} in the second triplet.
\textbf{JET}$^t$ only predicts 1 triplet due to the model's limitation {\color{black}(\textbf{JET}$^t$ is not able to handle the case of one target connecting to multiple opinion spans)}.
\textbf{JET}$^o$ is able to predict 2 triplets correctly.
In the second example, the gold data contains two triplets.
\citet{peng2019knowing} is able to correctly predict all the targets and opinion spans.
However, it incorrectly connects each target to both two opinion spans.
Our joint models \textbf{JET}$^t$ and \textbf{JET}$^o$ are both able to make the correct prediction.


\subsection{Ablation Study}

We also conduct an ablation study for \textbf{JET}$^t(M=6)_{\scriptscriptstyle \text{+ BERT}}$ and \textbf{JET}$^o(M=6)_{\scriptscriptstyle \text{+ BERT}}$ on dev set of the 4 datasets, presented in Table~\ref{tab:ablation}.
``$+$char embedding'' denotes concatenating character embedding into word representation.
The results show that concatenating character embedding mostly has no much positive impact on the performance, which we believe is due to data sparsity.
``$-$offset features'' denotes removing $f_r (j,k)$ in the feature score calculation, Equation \ref{eq:triplet}.
$F_1$ scores drop more on the \textbf{JET}$^t(M=6)_{\scriptscriptstyle \text{+ BERT}}$, this further confirms that modeling the opinion span is more difficult than target.
\textcolor{black}{``$-$opinion features'' denotes removing $f_o(\vec{g}_{i+j,i+k}) $ in the feature score calculation in Equation \ref{eq:triplet}.
$F_1$ scores drop consistently, implying the importance of such features for opinion spans.}

\begin{table}[t]
\centering
\scalebox{0.66}{
\begin{tabular}{lllll}
\toprule
\multicolumn{1}{l}{\multirow{2}{*}{Model}} & \multicolumn{2}{c}{\texttt{14Rest}} & \multicolumn{2}{c}{\texttt{14Lap}} \\ 
\multicolumn{1}{c}{} & \textbf{JET}$^t$    & \textbf{JET}$^o$   & \textbf{JET}$^t$     & \textbf{JET}$^o$    \\ 
\midrule
$M=6_{\scriptscriptstyle \text{+ BERT}}$ &
58.41 & 62.40 & 47.86 & 51.04  \\
$+$char embedding & 59.13 & 62.23 & 47.71 & 51.38 \\
$-$offset features &  55.36 & 61.24 & 44.16 & 49.58 \\
$-$opinion span features &57.93 & 62.04 & 47.66 & 50.48 \\ 

 & \multicolumn{2}{c}{\texttt{15Rest}} & \multicolumn{2}{c}{\texttt{16Rest}} \\
 \multicolumn{1}{c}{} & \textbf{JET}$^t$    & \textbf{JET}$^o$   & \textbf{JET}$^t$     & \textbf{JET}$^o$ \\
$M=6_{\scriptscriptstyle \text{+ BERT}}$ & 52.66 & 57.53 & 57.85 & 63.83 \\
$+$char embedding &  51.28 & 56.84 & 57.11 & 63.95 \\
$-$offset features &  48.74 & 53.68 & 52.83 & 61.72 \\
$-$opinion span features & 51.37 & 56.92 & 57.16 & 62.71 \\
\bottomrule
\end{tabular}
}
\vspace{-2mm}
\caption{Ablation Study ($F_1$)}
\label{tab:ablation}
\vspace{-6mm}
\end{table}


\subsection{Ensemble Analysis}
\begin{table}[t]
\centering
\scalebox{0.69}{
\begin{tabular}{llcccc}
\toprule
Dataset &  Model & $P.$ & $R.$ & $F_1$ \\ 
\midrule
\multirow{4}{*} {\texttt{14Rest}} 
&{\textbf{JET}$^{t}$} & 63.44 & 54.12 & 58.41\\
&{\textbf{JET}$^{o}$} & 70.56 & 55.94 & 62.40\\
\cdashline{2-5}
&\textbf{JET}$^{o\rightarrow t}$ &61.28 & 63.38& 62.31\\
&\textbf{JET}$^{t\rightarrow o}$ &61.10 & 63.98 & \bf{62.51}\\
\midrule
\multirow{4}{*} {\texttt{14Lap}} 
&{\textbf{JET}$^{t}$}&  53.53 & 43.28 & 47.86 \\
&{\textbf{JET}$^{o}$}&  55.39 & 47.33 & 51.04\\
\cdashline{2-5}
&\textbf{JET}$^{o\rightarrow  t}$&  48.68 & 51.01 & 49.82 \\
&\textbf{JET}$^{t\rightarrow  o}$& 49.57 & 53.22 & \bf{51.33} \\
\midrule
\multirow{4}{*}{ \texttt{15Rest}} 
&{\textbf{JET}$^{t}$}  &68.20 & 42.89	& 52.66\\
&{\textbf{JET}$^{o}$}  &64.45 & 51.96 & 57.53\\
\cdashline{2-5}
&\textbf{JET}$^{o\rightarrow  t}$  &61.41 & 53.81 & 57.36\\
&\textbf{JET}$^{t\rightarrow  o}$  &61.75 & 55.26 & \bf{58.32} \\
\midrule
\multirow{4}{*} {\texttt{16Rest}} 
&{\textbf{JET}$^{t}$} &65.28	& 51.95	& 57.85\\ 
&{\textbf{JET}$^{o}$} & 70.42 & 58.37 & \bf{63.83} \\ 
\cdashline{2-5}
&\textbf{JET}$^{o\rightarrow  t}$ &61.94 &62.06& 62.00\\ 
&\textbf{JET}$^{t\rightarrow  o}$ & 62.50 & 63.23 & 62.86\\ 
\bottomrule
\end{tabular} 
}
\vspace{-2.5mm}
\caption{Results for Ensemble{\color{black}. We use the models {\textbf{JET}$^{t}$} and {\textbf{JET}$^{o}$} (with BERT, $M=6$) as base models for building two ensemble models on 4 datasets.}}
\label{tab:ensemble}
\vspace{-5mm}
\end{table}
\label{sec:approach:ensemble}
{\color{black}As mentioned earlier, \textbf{JET}$^o$ is proposed to overcome the limitation of \textbf{JET}$^t$, and vice versa.
We believe that such two models complement each other.
Hence, we propose two ensemble models \textbf{JET}$^{o\rightarrow t}$ and \textbf{JET}$^{t\rightarrow o}$ to properly merge the results produced by \textbf{JET}$^t$ and \textbf{JET}$^o$.
\textbf{JET}$^{o\rightarrow t}$ merges results of \textbf{JET}$^{o}$ towards \textbf{JET}$^{t}$ by adding distinct triplets from \textbf{JET}$^{o}$ to \textbf{JET}$^{t}$, and analogously for \textbf{JET}$^{t\rightarrow o}$.}
\textcolor{black}{
We discuss how we build the ensemble models based on the two models \textbf{JET}$^{t}$ and \textbf{JET}$^{o}$ (with BERT, $M=6$).
First we call two triplets are \textit{overlap} with one another if two targets overlap {\em and} any of their opinions overlap with one another.
The ensemble model \textbf{JET}$^{o\rightarrow t}$ merges results from \textbf{JET}$^{o}$ towards \textbf{JET}$^{t}$.
Specifically, within the same instance, if a triplet produced by \textbf{JET}$^{o}$ does not {overlap} with any triplet produced by \textbf{JET}$^{t}$, we augment the prediction space with such an additional triplet.
After going through each triplet produced by \textbf{JET}$^{o}$, we regard the expanded predictions as the output of the ensemble model \textbf{JET}$^{o\rightarrow t}$.
Similarly, we merge the result from \textbf{JET}$^{t}$ towards \textbf{JET}$^{o}$ to obtain the result for the ensemble model \textbf{JET}$^{t\rightarrow o}$.}

\textcolor{black}{
We report results for ensemble models \textbf{JET}$^{o\rightarrow  t}$ and \textbf{JET}$^{t\rightarrow o}$ presented in Table~\ref{tab:ensemble}.   
As we can see, on {\texttt{14Rest}}, {\texttt{14Lap}} and {\texttt{15Rest}}, the ensemble model \textbf{JET}$^{t\rightarrow o}$ is able to achieve better $F_1$ score than \textbf{JET}$^t$ and \textbf{JET}$^o$. However, such a simple ensemble approach appears to be less effective on {\texttt{16Rest}}. It is worth highlighting that the ensemble models have significant improvements in terms of recall score. Note that the recall score reflects the number of gold triplets extracted. Such improvement confirms our earlier hypothesis that the two models largely complement each other.}

\section{Related Work}


ASTE is highly related to another research topic -- Aspect Based Sentiment Analysis (ABSA)~\cite{pontiki-EtAl:2014:SemEval,pontiki2016semeval}.
Such a research topic focuses on identifying aspect categories, recognizing aspect targets as well as the associated sentiment.
There exist a few tasks derived from ABSA.
Target extraction~\cite{chernyshevich-2014-ihs,san-vicente-etal-2015-elixa,yin2016unsupervised,lample2016neural,li2018aspect,ma-etal-2019-exploring} is a task that focuses on recognizing all the targets which are either aspect terms or named entities.
Such a task is mostly regarded as a sequence labeling problem solvable by CRF-based methods.
Aspect sentiment analysis or targeted sentiment analysis is another popular task.
Such a task either refers to predicting sentiment polarity for a given target~\cite{dong2014adaptive,chen-EtAl:2017:EMNLP20171,xue2018aspect,wang2018learning,P18-1088,P18-1087, peng2018learning, xu2020} or joint extraction of targets as well as sentiment associated with each target~\cite{mitchell2013open,zhang2015neural,li2017sentimentscope,ma2018joint,li-lu-2019-learning,li2019unified}.
The former mostly relies on different neural networks such as self-attention~\cite{liu2017EACL} or memory networks~\cite{tang2016aspect} to generate an opinion representation for a given target for further classification.
The latter mostly regards the task as a sequence labeling problem by applying CRF-based approaches.
Another related task -- target and opinion span co-extraction~\cite{qiu-etal-2011-opinion,P13-1172,P14-1030,D15-1168, wang2017coupled,P18-2094,dai2019neural} is also often regarded as a sequence labeling problem.

\section{Conclusion}
In this work, we propose a novel position-aware tagging scheme by enriching label expressiveness to address a limitation associated with existing works.
Such a tagging scheme is able to specify the connection among three elements -- a target, the target sentiment as well as an opinion span in an aspect sentiment triplet for the ASTE task.
Based on the position-aware tagging scheme, we propose a novel approach \textbf{JET} that is capable of jointly extracting the aspect sentiment triplets.
We also design factorized feature representations so as to effectively capture the interaction.
We conduct extensive experiments and results show that our models outperform strong baselines significantly with detailed analysis.
Future work includes finding applications of our novel tagging scheme in other  tasks involving extracting triplets as well as extending our approach to support other tasks within sentiment analysis.


\section*{Acknowledgements}

We would like to thank the anonymous reviewers for their helpful comments. 
This research is partially supported by Ministry of Education, Singapore, under its Academic Research Fund (AcRF) Tier 2 Programme (MOE AcRF Tier 2 Award No: MOE2017-T2-1-156). 
Any opinions, findings and conclusions or recommendations expressed in this material are those of the authors and do not reflect the views of the Ministry of Education, Singapore.

\bibliography{emnlp2020}
\bibliographystyle{acl_natbib}
\appendix
\section{More Data Statistics}
\begin{figure}[h]
\centering
\begin{subfigure}{0.45\linewidth}
   \centering
    \begin{adjustwidth}{-0.4cm}{0.0cm}
    \begin{tikzpicture}
    \pgfplotsset{width=5cm,compat=1.8}
    \begin{axis}
    [
                title={},
                legend style={font=\fontsize{6}{1}\selectfont},
                xticklabel style = {font=\fontsize{6}{1}\selectfont},
                yticklabel style = {font=\fontsize{6}{1}\selectfont},
                ylabel={}, 
                xmin=1, xmax=9,
                ymin=20, ymax=100,
                xtick={1,2,3,4,5,6,7,8,9,10},
                xticklabels = {1,2,3,4,5,6,7,8,9,$>=10$},
                ytick={20,30,40,50,60, 70, 80, 90, 100},
                legend pos= south east,
            ]
\addplot [mark=square, color=red] plot coordinates {
(1, 76.05) (2, 93.07) (3, 97.52) (4, 98.55) (5, 99.36)(6, 99.57)(7, 99.62) (8, 99.62) (9, 99.74) (10, 100)};
\addplot  [mark=diamond*, color=blue] plot coordinates {
(1, 88.92) (2, 96.19) (3, 98.20)  (4, 99.44) (5, 99.79) (6, 100)(7, 100) (8, 100) (9, 100) (10,100)
};
\addplot [mark=o, color=black] plot coordinates {
(1, 25.75) (2, 46.32) (3, 63.09) ( 4, 74.38) (5, 80.50) (6, 85.80)(7, 89.31) (8,91.79) (9, 93.93) (10,100)
 };
    \legend{target, opinion Span, offset}
    \end{axis}
    
    \end{tikzpicture}
   \vspace{-5mm}
    \end{adjustwidth}
    \centering
    \vspace{-3mm}
    \caption{{\texttt{14Rest}}}
\end{subfigure}
\hfill
\begin{subfigure}{0.45\linewidth}
   \centering
   \begin{adjustwidth}{-0.6cm}{0.0cm}
       
           \begin{tikzpicture}
            \pgfplotsset{width=5cm,compat=1.8}
            \begin{axis}
    [
                title={},
                legend style={font=\fontsize{6}{1}\selectfont},
                xticklabel style = {font=\fontsize{6}{1}\selectfont},
                yticklabel style = {font=\fontsize{6}{1}\selectfont},
                ylabel={}, 
                xmin=1, xmax=9,
                ymin=20, ymax=100,
                xtick={1,2,3,4,5,6,7,8,9,10},
                xticklabels = {1,2,3,4,5,6,7,8,9,$>=10$},
                ytick={20,30,40,50,60, 70, 80, 90, 100},
                legend pos= south east,
            ]
    \addplot [mark=square, color=red] plot coordinates {
    (1, 65.96) (2, 93.01) (3, 97.67) (4, 99.25) (5, 99.66)(6, 100)(7, 100) (8, 100) (9, 100) (10, 100)};
    \addplot  [mark=diamond*, color=blue]  plot coordinates {
    (1, 84.38) (2, 94.18) (3, 97.95)  (4, 99.45) (5, 99.86) (6, 99.93)(7, 100) (8, 100) (9, 100) (10,100)
    };
    \addplot[mark=o, color=black]  plot coordinates {
    (1, 21.58) (2, 41.99) (3, 56.51) ( 4, 67.26) (5, 75.21) (6, 81.37)(7, 85.07) (8,88.42) (9, 90.75) (10,100)
     };
    \legend{target, opinion Span, offset}
    \end{axis}
    
    \end{tikzpicture}
       \vspace{-5mm}
    \end{adjustwidth}
    \centering
    \vspace{-3mm}
   \caption{{\texttt{14Lap}}}
\end{subfigure}
\\
\begin{subfigure}{0.45\linewidth}
   \centering
   \begin{adjustwidth}{-0.4cm}{0.0cm}
       
           \begin{tikzpicture}
            \pgfplotsset{width=5cm,compat=1.8}
            \begin{axis}
    [
                title={},
                legend style={font=\fontsize{6}{1}\selectfont},
                xticklabel style = {font=\fontsize{6}{1}\selectfont},
                yticklabel style = {font=\fontsize{6}{1}\selectfont},
                ylabel={}, 
                xmin=1, xmax=9,
                ymin=20, ymax=100,
                xtick={1,2,3,4,5,6,7,8,9,10},
                xticklabels = {1,2,3,4,5,6,7,8,9,$>=10$},
                ytick={20,30,40,50,60, 70, 80, 90, 100},
                legend pos= south east,
            ]
        \addplot[mark=square, color=red] plot coordinates {
        (1, 74.53) (2, 91.61) (3, 95.95) (4, 97.83) (5, 99.31)(6, 99.61)(7, 99.80) (8, 99.90) (9, 100) (10, 100)};
        \addplot [mark=diamond*, color=blue] plot coordinates {
        (1, 89.44) (2, 96.74) (3, 98.42)  (4, 99.41) (5, 99.70) (6, 99.90)(7, 100) (8, 100) (9, 99.94) (10,100)
        };
        \addplot  [mark=o, color=black] plot coordinates {
        (1, 23.69) (2, 45.41) (3, 61.70) ( 4, 73.54) (5, 80.36) (6, 85.19)(7, 88.25) (8,91.61) (9, 93.68) (10,100)
         };
    \legend{target, opinion Span, offset}
    \end{axis}
    \end{tikzpicture}
       \vspace{-5mm}
    \end{adjustwidth}
    \centering
    \vspace{-3mm}
   \caption{{\texttt{15Rest}}}
\end{subfigure}
\hfill
\begin{subfigure}{0.45\linewidth}
   \centering
   \begin{adjustwidth}{-0.6cm}{0.0cm}
       
           \begin{tikzpicture}
            \pgfplotsset{width=5cm,compat=1.8}
            \begin{axis}
    [
                title={},
                legend style={font=\fontsize{6}{1}\selectfont},
                xticklabel style = {font=\fontsize{6}{1}\selectfont},
                yticklabel style = {font=\fontsize{6}{1}\selectfont},
                ylabel={}, 
                xmin=1, xmax=9,
                ymin=20, ymax=100,
                xtick={1,2,3,4,5,6,7,8,9,10},
                xticklabels = {1,2,3,4,5,6,7,8,9,$>=10$},
                ytick={20,30,40,50,60, 70, 80, 90, 100},
                legend pos= south east,
            ]
       \addplot[mark=square, color=red] plot coordinates {
        (1, 74.32) (2, 90.67) (3, 95.91) (4, 97.92) (5, 99.50)(6, 99.64)(7, 99.78) (8, 99.86) (9, 99.93) (10, 100)};
        \addplot  [mark=diamond*, color=blue] plot coordinates {
        (1, 88.24) (2, 96.20) (3, 98.57)  (4, 99.57) (5, 99.71) (6, 99.93)(7, 100) (8, 100) (9, 100) (10,100)
        };
        \addplot [mark=o, color=black] plot coordinates {
        (1, 24.53) (2, 45.98) (3, 62.55) ( 4, 74.46) (5, 81.35) (6, 85.65)(7, 89.81) (8,92.47) (9, 94.48) (10,100)
         };
    \legend{target, opinion Span, offset}
    \end{axis}
            
            \end{tikzpicture}
       \vspace{-5mm}
    \end{adjustwidth}
    \centering
    \vspace{-3mm}
   \caption{{\texttt{16Rest}}}
\end{subfigure}
\centering
\caption{Accumulative percentage ($y$-axis) in the training data of different lengths ($x$-axis) for targets, opinion spans and offsets on the 4 datasets.}
\label{fig:f1_diff_lengths_4datasets}
\end{figure}
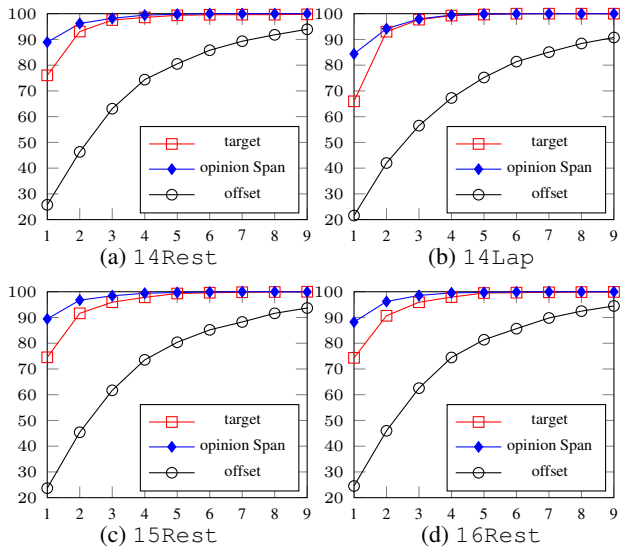

We present the statistics of accumulative percentage of different lengths for targets, opinion spans and offsets in the training data on 4 datasets {\texttt{14Rest}}, {\texttt{14Rest}}, {\texttt{15Rest}} and {\texttt{16Rest}} in Figure~\ref{fig:f1_diff_lengths_4datasets}.
As we mentioned in the main paper, similar patterns are observed on accumulative statistics on these 4 datasets.
\textcolor{black}{We also present the statistics of the number of targets with a single opinion span and with multiple opinion spans, and the number of opinion associated with a single target span and with multiple target spans, shown in Table \ref{tab:dataset_1}.}

\vspace{-3mm}

\section{Experimental Details}
We test our model on Intel(R) Xeon(R) Gold 6132 CPU, with PyTorch version 1.40. 
The average run time is 3300 sec/epoch, 1800 sec/epoch, 1170 sec/epoch, 1600 sec/epoch on  {\texttt{14Rest}}, {\texttt{14Rest}}, {\texttt{15Rest}} and {\texttt{16Rest}}  datasets respectively when $M=6$.
The total number of parameters is 2.5M.

\begin{table*}[!t]
\centering
\scalebox{0.68}{
\begin{tabular}{ll|cccc}
\toprule 
\multirow{2}{*}{\textbf{Dataset}} & &\# of Target with & \# of Target with &\# of Opinion with & \# of Opinion with \\
& &One Opinion Span & Multiple Opinion Spans & One Target Span & Multiple Target Spans \\\hline
\multirow{3}{*}{\texttt{14Rest}}
& Train & 1809 & 242 & 1893 & 193\\ 
& Dev & 433 & 67 & 444 & 59\\
& Test & 720 & 128 & 767 & 87\\\hline
\multirow{3}{*}{\texttt{14Lap}}
& Train & 1121 & 160 & 1114 & 154\\ 
& Dev & 252 & 44 & 270 & 34\\
& Test & 396 & 67 & 420 & 54\\\hline
\multirow{3}{*}{\texttt{15Rest}}
& Train & 734 & 128 & 893 & 48\\ 
& Dev & 180 & 33 & 224 & 12 \\
& Test & 385 & 47 & 438 & 23\\\hline
\multirow{3}{*}{\texttt{16Rest}}
& Train & 1029 & 169 & 1240 & 67\\ 
& Dev & 258 & 38 & 304 & 15\\
& Test & 396 & 56 & 452 & 23\\
\bottomrule
\end{tabular}
}
\caption{Statistics of 4 datasets.}
\label{tab:dataset_1}
\vspace{-4mm}
\end{table*}

For hyper-parameter, we use pre-trained 300d  GloVe~\cite{pennington2014glove} to initialize the word embeddings. 
We use 100 as the embedding size of $\vec{w_r}$ (offset embedding).
For out-of-vocabulary words as well as $\vec{w_r}$, we randomly sample their embeddings from the uniform distribution $\mathcal{U}(-0.1, 0.1)$, as done in~\cite{kim:2014:EMNLP2014}.
We use the bi-directional LSTM with the hidden size $300$.
We train our model for a maximal of 20 epochs using Adam~\cite{kingma2014adam} as the optimizer with batch size $1$ and dropout rate $0.5$ for datasets in restaurant domain and $0.7$ for laptop domain. We manually tune the dropout rate from 0.4 to 0.7,
and select the best model parameters based on the best $F_1$ score on the development data and apply it to the test data for evaluation.
For experiments with contextualised representation, we adopt the pre-trained language model BERT \cite{devlin2019bert}.  Specifically, we use bert-as-service \cite{xiao2018bertservice} to generate the contextualized word embedding without fine-tuning. We use the representation from the last layer of the uncased version of BERT base model for our experiments.

\section{Experimental Results}
Table \ref{tab:main_results_1} presents the experimental result on the previous released dataset by \cite{peng2019knowing}.

\begin{table*}[!t]

    \centering
    \scalebox{0.62}{
    \begin{tabular}{lcccc|cccc|cccc|cccc}
    \toprule
      \multirow{2}{*}{\textbf{Models}} & \multicolumn{4}{c}{\texttt{14Rest}} & \multicolumn{4}{c}{\texttt{14Lap}} & \multicolumn{4}{c}{\texttt{15Rest}}  & \multicolumn{4}{c}{\texttt{16Rest}}    \\ 
      &Dev $F_1$ & $P.$ & $R.$ & $F_1$&Dev $F_1$ & $P.$ & $R.$ & $F_1$&Dev $F_1$ & $P.$ & $R.$ & $F_1$ &Dev $F_1$ & $P.$ & $R.$ & $F_1$ \\ \midrule
       
        \textbf{CMLA+} & -& 40.11 & 46.63 & 43.12 &-&  31.40 & 34.60 & 32.90 &-&  34.40 & 37.60 & 35.90 &-&  43.60 & 39.80 & 41.60          \\
        \textbf{RINANTE+} &-&  31.07 & 37.63 & 34.03 &-&  23.10 & 17.60 & 20.00  &-&  29.40 & 26.90 & 28.00 &-&  27.10 & 20.50 & 23.30  \\
        \textbf{Li-unified-R} &-&  41.44 & 68.79 & 51.68 &-&  42.25 & 42.78 & 42.47 &-&  43.34 & 50.73 & 46.69 &-&  38.19 & 53.47 & 44.51 \\
        \citet{peng2019knowing} &-&  44.18 & 62.99 & 51.89 &-&  40.40 & 47.24 & 43.50 &-&  40.97 & 54.68 & 46.79 &-&  46.76 & 62.97 & 53.62 \\ 
        \midrule
        \textbf{JET}$^t$ $(M=2)$& 47.06& 70.00 & 34.92& 46.59& 35.00& 63.69& 23.27 & 34.08 &47.13&  64.80 & 27.91 & 39.02&42.32& 70.76 & 35.91 & 47.65 \\
        \textbf{JET}$^t$ $(M=3)$&56.15 & 73.15 & 43.62& 54.65& 43.72& 54.18 & 30.41 & 38.95 &53.23 & 66.52 & 33.19 & 44.28&50.50& 66.35 & 44.95 & 53.59\\
        \textbf{JET}$^t$ $(M=4)$ &57.47&  70.25 &49.30 & 57.94 &43.19& 57.46& 31.43 & 40.63 &58.05& 64.77 & 42.42& 51.26& 53.57&  68.79& 48.82& 57.11 \\
        \textbf{JET}$^t$ $(M=5)$&59.15 & 66.20 & 49.77&56.82 &45.47& 59.50 & 33.88 & 43.17 &59.37 &64.14 & 40.88 &49.93 &54.16& 66.86& 50.32 &57.42\\
        \textbf{JET}$^t$ $(M=6)$&\underline{59.51}& 70.39&51.86& \bf{59.72}&\underline{45.83}& 57.98& 36.33 &\bf{44.67} &\underline{60.00}& 61.99&43.74& \bf{51.29}& \underline{55.88} &68.99&51.18&\bf{58.77}\\[1.5mm]
        \textbf{JET}$^o$ $(M=2)$&45.02&  66.30 & 35.38& 46.14 &33.01&  50.43 & 23.88 & 32.41 &46.80&  58.88 & 25.49 & 35.58&40.33& 60.47 & 39.14 & 47.52 \\        
        \textbf{JET}$^o$ $(M=3)$&53.14&  62.31 & 43.16& 50.99 &38.99& 55.37 & 33.67 & 41.88&54.59&  55.99 & 38.02 & 45.29 &47.87& 69.45 & 46.45 & 55.67\\
        \textbf{JET}$^o$ $(M=4)$&58.19& 63.84 & 52.44 & 57.58 &40.87& 49.86& 36.33& 42.03 &57.14&  57.57 & 42.64& 48.99&53.99& 73.98 &54.41 & 62.70\\  
        \textbf{JET}$^o$ $(M=5)$&57.94&  64.31 &	54.99 &	59.29 &\underline{43.23}& 52.36& 40.82& \bf{45.87} &59.51&  52.02 & 48.13 & {50.00} &\underline{56.08}& 66.91 & 58.71 & \bf{62.54}\\
        \textbf{JET}$^o$ $(M=6)$&\underline{58.66}& 62.26&	56.84&\bf{59.43}& 42.50&52.01	&39.59&	44.96 &\underline{60.32}&63.25&	46.15&	\bf{53.37}&55.63& 66.58 & 57.85 & 61.91\\ [1.5mm] 
        \multicolumn{4}{l}{\textbf{+ Contextualized Word Representation (BERT)}}& & &&&&&&&&&&&\\
        \textbf{JET}$^t$ $(M=6)_{\scriptscriptstyle \text{+ BERT}}$& 61.01&  70.20	& 53.02& 60.41 & 49.07 & 51.48 & 42.65 & 46.65 & 62.96 & 62.14 & 47.25	& 53.68 & 60.41 & 71.12	& 57.20	& 63.41 \\
        \textbf{JET}$^o$ $(M=6)_{\scriptscriptstyle \text{+ BERT}}$ & 60.86 & 67.97	& 60.32	& 63.92 & 45.76 & 58.47	& 43.67	& 50.00  & 64.12 & 58.35 & 51.43 & 54.67 & 60.17 & 64.77 & 61.29 & 62.98 \\
        
  \bottomrule
    \end{tabular}
    }
    \caption{The experimental results on the previous released datasets ASTE-Data-V1. The underlined scores indicate the best results on the dev set, and the highlighted scores are the corresponding test  results. }
    \label{tab:main_results_1}
    \vspace{-4mm}
\end{table*}

\section{Decoding based on Viterbi}
Let $\mathcal{T}={ \{ B^{\epsilon}_{j,k}, S^{\epsilon}_{j,k}, I, E, O} \}$ as the new tag set under our position-aware tagging scheme, where $\epsilon$ denotes the sentiment polarity for the target, and $j,k$  indicate the position information which are the distances between the two ends of an opinion span and the starting position of a target respectively.

As we know, $|j| \leq |k| \leq M$, $\epsilon \in \{+, 0, -\}$.

$$O(|\mathcal{T}|)=O(|\epsilon| M^2)=O(M^2)$$ 

We define the sub-tags of $B^{\epsilon}_{j,k}, S^{\epsilon}_{j,k}$ as $B$ and $S$ respectively, and the sub-tags of $I,O,E$ as themselves.
We use the bar on top to denote the sub-tag.
For example, $\bar{u}$ is the subtag of $u \in \mathcal{T}$.



We use $\pi(i,v)$ to denote the score for the optimal sequence $\{ \vec{y}^{*}_1 \cdots \vec{y}^{*}_i \}$ among all the possible sequences whose last tag is $v$.

Given the input $\vec{x}$ of length $n$, we aim to obtain the optimal sequence $\vec{y}^{*} = \{ \vec{y}^{*}_1 \cdots \vec{y}^{*}_n \} $. \\
\begin{itemize}
    \item Base Case for all the $v \in \mathcal{T}$ \\
    If $v \in \{ I,E,O \}$:
    $$\pi(1, v) =  \psi_{START, \bar{v}} + f_t(\vec{h}_1)_{\bar{\vec{v}}}$$
    
    If $v \in \{ B^{\epsilon}_{j,k}, S^{\epsilon}_{j,k}\}$:
    \setlength{\abovedisplayskip}{4pt} \setlength{\abovedisplayshortskip}{4pt}
    \setlength{\belowdisplayskip}{4pt} \setlength{\belowdisplayshortskip}{4pt}
    \begin{align}
        \pi(1, v) &=  \psi_{START, \bar{v}} + \Phi_{v}(\vec{x},1)  \nonumber\\
          &=  \psi_{START, \bar{v}} + f_t(\vec{h}_1)_{\bar{\vec{v}}} \nonumber\\& + f_s([\vec{g}_{1+j,1+k};\overleftarrow{\vec{h}_{1}}])_{\epsilon} + f_o(\vec{g}_{1+j,1+k}) \nonumber\\& + f_r(j,k) \nonumber
    \end{align}
    where  $f_t(\vec{h}_i)_{\bar{\vec{v}}}$, $f_s([\vec{g}_{1+j,1+k};\overleftarrow{\vec{h}_{1}}])_{\epsilon}$, $f_o(\vec{g}_{1+j,1+k})$, and $f_r(j,k)$ are the factorized feature score  mentioned in the section 2.2.2.
    \item Loop forward for $i \in \{2, \cdots ,n \}$ and all the $v \in \mathcal{T}$ \\
    If $v \in \{ I,E,O \}$:
    \setlength{\abovedisplayskip}{4pt} \setlength{\abovedisplayshortskip}{4pt}
    \setlength{\belowdisplayskip}{4pt} \setlength{\belowdisplayshortskip}{4pt}
    \begin{align}
        \pi(i, v) &=  \max_{u \in \mathcal{T}} \{ \pi(i-1, u) + \psi_{\bar{u}, \bar{v}}  + f_t(\vec{h}_i)_{\bar{\vec{v}}} \} \nonumber
    \end{align}
    If $v \in \{ B^{\epsilon}_{j,k}, S^{\epsilon}_{j,k}\}$:
    \setlength{\abovedisplayskip}{4pt} \setlength{\abovedisplayshortskip}{4pt}
    \setlength{\belowdisplayskip}{4pt} \setlength{\belowdisplayshortskip}{4pt}
    \begin{align}
        \pi(i, v) & = \max_{u \in \mathcal{T}} \{ \pi(i-1,u) + \psi_{\bar{u}, \bar{v}} + \Phi_{v}(\vec{x},i) \} \nonumber\\
       & = \max_{({u} \in \mathcal{T}; \; j,k \in [-M,M] ; \; \epsilon \in \{+,0,-\})} \{ \nonumber \\ &
       \pi(i-1,{u}) + \psi_{\bar{u}, \bar{v}} +   f_t(\vec{h}_i)_{\bar{\vec{v}}} \nonumber\\ & + f_s([\vec{g}_{i+j,i+k};\overleftarrow{\vec{h}_{i}}])_{\epsilon} + f_o(\vec{g}_{i+j,i+k})  \nonumber \\&
       + f_r(j,k) \}
       \nonumber
    \end{align}
    \item Backtrack for the optimal sequence $\vec{y}^{*} = \{ \vec{y}^{*}_1 \cdots \vec{y}^{*}_n \}$ \\
    \setlength{\abovedisplayskip}{4pt} \setlength{\abovedisplayshortskip}{4pt}
    \setlength{\belowdisplayskip}{4pt} \setlength{\belowdisplayshortskip}{4pt}
    $$\vec{y}^{*}_n = \argmax_{v \in \mathcal{T}} {\{ \pi(n,v) + \psi_{\bar{v}, STOP} \}}$$
    Loop for $i \in \{n - 1, \cdots ,1 \}$ 
    $$\vec{y}^{*}_i = \argmax_{v \in \mathcal{T}} {\{ \pi(i,v) + \psi_{\bar{v}, \bar{\vec{y}}^{*}_{i+1}} \}}$$
\end{itemize}
Note that $START$ appears before the start of the input sentence and $STOP$ appears after the end of the input sentence.

The time complexity is $O(n|\mathcal{T}|)=O(nM^2)$.

\section{Analysis}
\subsection{Robustness Analysis}
\begin{figure}[t!]
\centering
\begin{subfigure}{0.45\linewidth}
   \centering
    \begin{adjustwidth}{-0.4cm}{0.0cm}
    \begin{tikzpicture}
    \pgfplotsset{width=5cm,compat=1.9}
    \begin{axis}
    [
                title={},
                legend style={font=\fontsize{5}{1}\selectfont},
                legend style={at={(0.5,0.85)},anchor=south,legend columns=2}, 
                xticklabel style = {font=\fontsize{6}{1}\selectfont},
                yticklabel style = {font=\fontsize{6}{1}\selectfont},
                xmin=1, xmax=6,
                ymin=0, ymax=115,
                xtick={1,2,3,4,5,6,7},
                xticklabels = {1,2,3,4,5,6,$>=7$},
                ytick={0.0,20,40,60, 80,100},
            ]
    \addplot [mark=square, color=red] plot coordinates {
        (1,52.82) (2, 43.51) (3, 35.29) (4, 0) (5, 0)(6, 0)};
        \addplot [mark=diamond*, color=blue] plot coordinates {
        (1, 51.90) (2, 21.62) (3,11.11)  (4,0) (5,0) (6, 0) };
        \addplot[mark=o, color=black] plot coordinates {
        (1, 61.24) (2, 67.01) (3, 50.31) ( 4, 36.76) (5, 40) (6, 28.07)};
    \legend{target \\opinion span\\ offset\\}
    \end{axis}
    \end{tikzpicture}
    \vspace{-5mm}
    \end{adjustwidth}
    \centering
    \vspace{-2mm}
    \caption{\textbf{JET}$^t (M=6)$}
\end{subfigure}
\hfill
\begin{subfigure}{0.45\linewidth}
   \centering
   \begin{adjustwidth}{-0.6cm}{0.0cm}
        \begin{tikzpicture}
        \pgfplotsset{width=5cm,compat=1.9}
        \begin{axis}
        [
                    title={},
                    legend style={font=\fontsize{5}{1}\selectfont}, 
                    legend style={at={(0.5,0.85)},anchor=south,legend columns=2}, 
                    xticklabel style = {font=\fontsize{6}{1}\selectfont},
                    yticklabel style = {font=\fontsize{6}{1}\selectfont},
                    xmin=1, xmax=6,
                    ymin=0, ymax=115,
                    xtick={1,2,3,4,5,6},
                    xticklabels = {1,2,3,4,5,6},
                    ytick={0.0,20,40,60, 80,100},
                ]
        \addplot [mark=square, color=red] plot coordinates {
        (1,55.57) (2, 46.65) (3, 41.51) (4, 15.38) (5, 0)(6, 0)(7, 0)};
        \addplot [mark=diamond*, color=blue] plot coordinates {
        (1, 53.85) (2, 35.14) (3, 18.75)  (4, 0) (5,0) (6, 0)(7, 0) };
        \addplot[mark=o, color=black] plot coordinates {
        (1, 64.37) (2, 68.20) (3, 55.03) ( 4, 47.62) (5, 38.55) (6, 26.42)};
        \legend{target \\opinion span\\ offset\\}
        \end{axis}
        \end{tikzpicture}
        \vspace{-5mm}
    \end{adjustwidth}
    \centering
    \vspace{-2mm}
   \caption{\textbf{JET}$^o (M=6)$}
   
\end{subfigure}
\centering
\caption{$F_1(\%)$ scores ($y$-axis) of different lengths ($x$-axis) for targets, opinion spans and offsets on the dataset {\texttt{14Lap}}.}
\label{fig:f1_diff_lengths_14lap}
\end{figure}
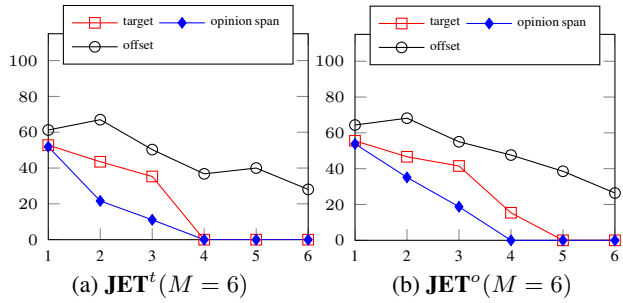

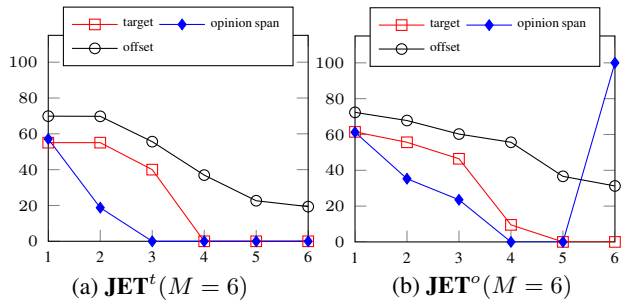
\begin{figure}[t!]
\centering
\begin{subfigure}{0.45\linewidth}
   \centering
    \begin{adjustwidth}{-0.4cm}{0.0cm}
    \begin{tikzpicture}
    \pgfplotsset{width=5cm,compat=1.9}
    \begin{axis}
    [
                title={},
                legend style={font=\fontsize{5}{1}\selectfont},
                legend style={at={(0.5,0.85)},anchor=south,legend columns=2}, 
                xticklabel style = {font=\fontsize{6}{1}\selectfont},
                yticklabel style = {font=\fontsize{6}{1}\selectfont},
                xmin=1, xmax=6,
                ymin=0, ymax=115,
                xtick={1,2,3,4,5,6,7},
                xticklabels = {1,2,3,4,5,6,$>=7$},
                ytick={0.0,20,40,60, 80,100},
            ]
        \addplot [mark=square, color=red] plot coordinates {
        (1,55.08) (2, 55.08) (3, 40) (4, 0) (5, 0)(6, 0)};
        \addplot [mark=diamond*, color=blue] plot coordinates {
        (1, 57.14) (2, 18.75) (3,0)  (4,0) (5,0) (6, 0) };
        \addplot[mark=o, color=black] plot coordinates {
        (1, 69.87) (2, 69.77) (3, 55.63) ( 4, 36.96) (5, 22.64) (6, 19.35)};
    \legend{target \\opinion span\\ offset\\}
    \end{axis}
    \end{tikzpicture}
    \vspace{-5mm}
    \end{adjustwidth}
    \centering
    \vspace{-2mm}
    \caption{\textbf{JET}$^t (M=6)$}
\end{subfigure}
\hfill
\begin{subfigure}{0.45\linewidth}
   \centering
   \begin{adjustwidth}{-0.6cm}{0.0cm}
        \begin{tikzpicture}
        \pgfplotsset{width=5cm,compat=1.9}
        \begin{axis}
        [
                    title={},
                    legend style={font=\fontsize{5}{1}\selectfont}, 
                    legend style={at={(0.5,0.85)},anchor=south,legend columns=2}, 
                    xticklabel style = {font=\fontsize{6}{1}\selectfont},
                    yticklabel style = {font=\fontsize{6}{1}\selectfont},
                    xmin=1, xmax=6,
                    ymin=0, ymax=115,
                    xtick={1,2,3,4,5,6},
                    xticklabels = {1,2,3,4,5,6},
                    ytick={0.0,20,40,60, 80,100},
                ]
        \addplot [mark=square, color=red] plot coordinates {
        (1, 61.44) (2, 55.63) (3, 46.43) (4, 9.52) (5, 0)(6, 0)};
        \addplot [mark=diamond*, color=blue] plot coordinates {
        (1, 61.25) (2, 35.29) (3, 23.53)  (4,0) (5, 0) (6, 100) };
        \addplot[mark=o, color=black] plot coordinates {
        (1, 72.34) (2, 67.78) (3, 60.23) ( 4, 55.65) (5, 36.62) (6, 31.25)};
        \legend{target \\opinion span\\ offset\\}
        \end{axis}
        \end{tikzpicture}
        \vspace{-5mm}
    \end{adjustwidth}
    \centering
    \vspace{-2mm}
   \caption{\textbf{JET}$^o (M=6)$}
\end{subfigure}
\centering
\caption{$F_1(\%)$ scores ($y$-axis) of different lengths ($x$-axis) for targets, opinion spans and offsets on the dataset {\texttt{15Rest}}.}
\label{fig:f1_diff_lengths_15rest}
\end{figure}

\begin{figure}[t!]
\centering
\begin{subfigure}{0.45\linewidth}
   \centering
    \begin{adjustwidth}{-0.4cm}{0.0cm}
    \begin{tikzpicture}
    \pgfplotsset{width=5cm,compat=1.9}
    \begin{axis}
    [
                title={},
                legend style={font=\fontsize{5}{1}\selectfont},
                legend style={at={(0.5,0.85)},anchor=south,legend columns=2}, 
                xticklabel style = {font=\fontsize{6}{1}\selectfont},
                yticklabel style = {font=\fontsize{6}{1}\selectfont},
                xmin=1, xmax=6,
                ymin=0, ymax=115,
                xtick={1,2,3,4,5,6,7},
                xticklabels = {1,2,3,4,5,6,$>=7$},
                ytick={0.0,20,40,60, 80,100},
            ]
    \addplot [mark=square, color=red] plot coordinates {
    (1,60.34) (2, 57.89) (3, 41.02) (4, 36.36) (5, 33.33)(6, 0)(7, 0)};
    \addplot [mark=diamond*, color=blue] plot coordinates {
    (1, 61.87) (2, 21.74) (3, 0)  (4, 25) (5,0) (6, 0)(7, 0) };
    \addplot [mark=o, color=black] plot coordinates {
    (1, 72.31) (2, 69.95) (3, 67.07) ( 4, 43.01) (5, 20.00) (6, 22.22)(7, 0) };
    \legend{target \\opinion span\\ offset\\}
    \end{axis}
    \end{tikzpicture}
    \end{adjustwidth}
    \centering
    \vspace{-3mm}
    \caption{\textbf{JET}$^t (M=6)$}
    \vspace{-2mm}
\end{subfigure}
\hfill
\begin{subfigure}{0.45\linewidth}
   \centering
   \begin{adjustwidth}{-0.6cm}{0.0cm}
        \begin{tikzpicture}
        \pgfplotsset{width=5cm,compat=1.9}
        \begin{axis}
        [
                    title={},
                    legend style={font=\fontsize{5}{1}\selectfont}, 
                    legend style={at={(0.5,0.85)},anchor=south,legend columns=2}, 
                    xticklabel style = {font=\fontsize{6}{1}\selectfont},
                    yticklabel style = {font=\fontsize{6}{1}\selectfont},
                    xmin=1, xmax=6,
                    ymin=0, ymax=115,
                    xtick={1,2,3,4,5,6},
                    xticklabels = {1,2,3,4,5,6},
                    ytick={0.0,20,40,60, 80,100},
                ]
        \addplot [mark=square, color=red] plot coordinates {
        (1,66.95) (2, 66.67) (3, 36.36) (4, 30.77) (5, 0)(6, 0)};
        \addplot [mark=diamond*, color=blue] plot coordinates {
        (1, 66.98) (2, 37.74) (3, 28.57)  (4, 36.36) (5,0) (6, 0) };
        \addplot[mark=o, color=black] plot coordinates {
        (1, 74.27) (2, 70.94) (3, 70.06) ( 4, 61.54) (5, 54.84) (6, 44.44)};
        \legend{target \\opinion span\\ offset\\}
        \end{axis}
        \end{tikzpicture}
        
    \end{adjustwidth}
    \centering
    \vspace{-3mm}
   \caption{\textbf{JET}$^o (M=6)$}
  \vspace{-2mm}
\end{subfigure}
\centering
\caption{$F_1(\%)$ scores ($y$-axis) of different lengths ($x$-axis) for targets, opinion spans and offsets on the dataset {\texttt{16Rest}}.}
\label{fig:f1_diff_lengths_16rest}
\vspace{-2mm}
\end{figure}
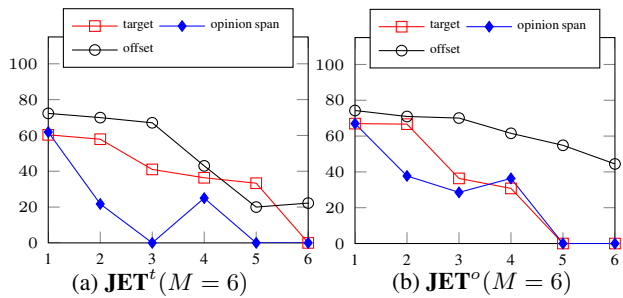

We present the performance on targets, opinion spans and offsets of different lengths for two models \textbf{JET}$^t (M=6)$ and \textbf{JET}$^o(M=6)$ with BERT on 3 datasets {\texttt{14Lap}},{\texttt{15Rest}} and {\texttt{16Rest}} in Figure~\ref{fig:f1_diff_lengths_14lap}, Figure~\ref{fig:f1_diff_lengths_15rest} and Figure~\ref{fig:f1_diff_lengths_16rest} respectively.
\subsection{Qualitative Analysis}

We present one additional example sentence selected from the test data as well as predictions by~\citet{peng2019knowing}, \textbf{JET}$^t$ and \textbf{JET}$^o$ in Table~\ref{tab:qualitative2}.
As we can see, the gold data contains two triplets.
\citet{peng2019knowing} only predicts 1 opinion span, and therefore incorrectly assigns the opinion span ``Good'' to the target ``price''.
\textbf{JET}$^t$ is able to make the correct predictions.
\textbf{JET}$^o$ only predicts 1 triplet correctly.
The qualitative analysis helps us to better understand the differences among these models.

\section{More Related Work}
The task of joint entity and relation extraction is also related to joint triplet extraction.
Different from our task, such a relation extraction task aims to extract a pair of entities (instead of a target and an opinion span) and their relation as a triplet in a joint manner.
\citet{miwa-sasaki-2014-modeling} and \citet{li-ji-2014-incremental} used approaches motivated by a table-filling method to jointly extract entity pairs as well as their relations.
The tree-structured neural networks~\cite{miwa-bansal-2016-end} and CRF-based approaches~\cite{adel-schutze-2017-global} were also adopted to capture rich context information for triplet extraction.
Recently, \citet{bekoulis-etal-2018-adversarial} used adversarial training~\cite{goodfellow15} for this task and results show that it performs more robustly in different domains.
Although these approaches may not be applied to our task ASTE, they may provide inspirations for future work.

\begin{table*}[t!]
     \centering
     \scalebox{0.8}{
     \begin{tabular}{ c  c c c   }
     \toprule
      Gold & \citet{peng2019knowing} & \textbf{JET}$^t$ & \textbf{JET}$^o$ \\ 
    \midrule
            \scalebox{0.7}{
     	    	\begin{tikzpicture}[node distance=1.0mm and 1.0mm, >=Stealth, 
			wordnode/.style={draw=none, minimum height=2mm, inner sep=0pt},
			chainLine/.style={line width=1pt,->, color=blue},
			opinionbox/.style={draw=black, rounded corners, fill=yellow!20, dashed},
			targetbox/.style={draw=black, rounded corners, fill=red!20},
			]

			\matrix (sent1) [matrix of nodes, ampersand replacement=\&, nodes in empty cells, execute at empty cell=\node{\strut};]
			{
				Good \& [1mm] \textbf{food} \&  at \&  the \&  right \&  [1mm] price \& ,  \\
			};

			\begin{pgfonlayer}{background}
			
			\node [targetbox, above=of sent1-1-2, yshift=-7mm, text height=-2mm, minimum height = 10mm, minimum width=10mm] (e1)  [] {\color{blue!80}\textbf{\textsc{0}}};	
			\node [opinionbox, above=of sent1-1-1, xshift=0mm, yshift=-6mm, text height=-5mm, minimum height = 7mm, minimum width=10mm] (o1)  [] {\color{blue!80}\textbf{\textsc{}}};
			
			\draw[chainLine,->]   (e1) to[out=150,in=45] (o1);

			\node [targetbox, above=of sent1-1-6, xshift=0mm, yshift=-7mm, text height=-2mm, minimum height = 10mm, minimum width=10mm] (e2)  [] {\color{blue!80}\textbf{\textsc{+}}};			
			\node [opinionbox, above=of sent1-1-5, xshift=0mm, yshift=-6mm, text height=-5mm, minimum height = 7mm, minimum width=9mm] (o2)  [] {\color{blue!80}\textbf{\textsc{}}};		
			
			\draw[chainLine,->]  (e2) to[out=150,in=45] (o2);
				
			\end{pgfonlayer}
			
			\end{tikzpicture} 
			}

      & 
      
       \scalebox{0.7}{
     	    	\begin{tikzpicture}[node distance=1.0mm and 1.0mm, >=Stealth, 
			wordnode/.style={draw=none, minimum height=2mm, inner sep=0pt},
			chainLine/.style={line width=1pt,->, color=blue},
			opinionbox/.style={draw=black, rounded corners, fill=yellow!20, dashed},
			targetbox/.style={draw=black, rounded corners, fill=red!20},
			]

			\matrix (sent1) [matrix of nodes, ampersand replacement=\&, nodes in empty cells, execute at empty cell=\node{\strut};]
			{
				Good \&  [1mm] \textbf{food} \&  at \&  the \&  right \&  [1mm] price \& ,  \\
			};

			\begin{pgfonlayer}{background}
			
			\node [targetbox, above=of sent1-1-2, yshift=-7mm, text height=-2mm, minimum height = 10mm, minimum width=10mm] (e1)  [] {\color{blue!80}\textbf{\textsc{0}}};	
			\node [opinionbox, above=of sent1-1-1, xshift=0mm, yshift=-6mm, text height=-5mm, minimum height = 7mm, minimum width=10mm] (o1)  [] {\color{blue!80}\textbf{\textsc{}}};
			
			\draw[chainLine,->]   (e1) to[out=150,in=45] (o1);

			\node [targetbox, above=of sent1-1-6, xshift=0mm, yshift=-7mm, text height=-2mm, minimum height = 10mm, minimum width=10mm] (e2)  [] {\color{blue!80}\textbf{\textsc{+}}};			
			
			\draw[chainLine,->]  (e2) to[out=150,in=50] (o1);
				
			\end{pgfonlayer}
			
			\end{tikzpicture} 
			}
     
      & 
      
       \scalebox{0.7}{
     	    	\begin{tikzpicture}[node distance=1.0mm and 1.0mm, >=Stealth, 
			wordnode/.style={draw=none, minimum height=2mm, inner sep=0pt},
			chainLine/.style={line width=1pt,->, color=blue},
			opinionbox/.style={draw=black, rounded corners, fill=yellow!20, dashed},
			targetbox/.style={draw=black, rounded corners, fill=red!20},
			]

			\matrix (sent1) [matrix of nodes, ampersand replacement=\&, nodes in empty cells, execute at empty cell=\node{\strut};]
			{
				Good \&  [1mm] \textbf{food} \&  at \&  the \&  right \&  [1mm]  price \& ,  \\
			};

			\begin{pgfonlayer}{background}
			
			\node [targetbox, above=of sent1-1-2, yshift=-7mm, text height=-2mm, minimum height = 10mm, minimum width=10mm] (e1)  [] {\color{blue!80}\textbf{\textsc{0}}};	
			\node [opinionbox, above=of sent1-1-1, xshift=0mm, yshift=-6mm, text height=-5mm, minimum height = 7mm, minimum width=10mm] (o1)  [] {\color{blue!80}\textbf{\textsc{}}};
			
			\draw[chainLine,->]   (e1) to[out=150,in=45] (o1);

			\node [targetbox, above=of sent1-1-6, xshift=0mm, yshift=-7mm, text height=-2mm, minimum height = 10mm, minimum width=10mm] (e2)  [] {\color{blue!80}\textbf{\textsc{+}}};			
			\node [opinionbox, above=of sent1-1-5, xshift=0mm, yshift=-6mm, text height=-5mm, minimum height = 7mm, minimum width=9mm] (o2)  [] {\color{blue!80}\textbf{\textsc{}}};		
			
			\draw[chainLine,->]  (e2) to[out=150,in=45] (o2);
				
			\end{pgfonlayer}
			
			\end{tikzpicture} 
			}
    
     &

 \scalebox{0.7}{
     	    	\begin{tikzpicture}[node distance=1.0mm and 1.0mm, >=Stealth, 
			wordnode/.style={draw=none, minimum height=2mm, inner sep=0pt},
			chainLine/.style={line width=1pt,->, color=blue},
			opinionbox/.style={draw=black, rounded corners, fill=yellow!20, dashed},
			targetbox/.style={draw=black, rounded corners, fill=red!20},
			]

			\matrix (sent1) [matrix of nodes, ampersand replacement=\&, nodes in empty cells, execute at empty cell=\node{\strut};]
			{
				Good \&  [1mm] \textbf{food} \&  at \&  the \&  right \&  [1mm] price \& ,  \\
			};

			\begin{pgfonlayer}{background}
			
			\node [targetbox, above=of sent1-1-2, yshift=-7mm, text height=-2mm, minimum height = 10mm, minimum width=10mm] (e1)  [] {\color{blue!80}\textbf{\textsc{0}}};	
			\node [opinionbox, above=of sent1-1-1, xshift=0mm, yshift=-6mm, text height=-5mm, minimum height = 7mm, minimum width=10mm] (o1)  [] {\color{blue!80}\textbf{\textsc{}}};
			
			\draw[chainLine,->]   (e1) to[out=150,in=45] (o1);

			
				
			\end{pgfonlayer}
			
			\end{tikzpicture} 
			}
      \\

      \bottomrule
      \end{tabular}
      }
      \centering
      \caption{Qualitative Analysis}
      \label{tab:qualitative2}
\end{table*}

\begin{algorithm}[H]
\SetAlgoLined
Initialization\;
\For{$i=1$}{
\For{ $ \bar{v} \in \{ I,E,O \}$}
{ $v = \bar{{v}}$\; \\
$\pi(1, v) = \psi_{START, \bar{v}} + f_t(\vec{h}_1)_{\bar{\vec{v}}}$}
\For{ $\bar{v} \in \{ B, S \} $}{
    \For{$j \in [-M, M]$}{
        \For{$k \in [j, M]$}{
            \For{$\epsilon \in \{+, 0, -\}$}{
            $v = \bar{{v}}_{j,k}^{\epsilon} $\;\\
            {$\pi(1, v) = \psi_{START, \bar{v}} +  f_t(\vec{h}_1)_{\bar{\vec{v}}} + f_s([\vec{g}_{1+j,1+k};\overleftarrow{\vec{h}_{1}}])_{\epsilon} + f_o(\vec{g}_{1+j,1+k})  + f_r(j,k) $}}}}}
}
Loop Forward\;
\For{$i \in \{2, \cdots ,n \}$}{
\For{ $ \bar{v} \in \{ I,E,O \}$}
{$v = \bar{{v}}$\; \\
$\pi(i, v) =  \max_{u \in \mathcal{T}} \{ {\pi(i-1, u) + \psi_{\bar{u}, \bar{v}}  + f_t(\vec{h}_i)_{\bar{\vec{v}}}} \}$ \\
}

\For{ $\bar{v} \in \{ B, S \} $}{
    \For{$j \in [-M, M]$}{
        \For{$k \in [j, M]$}{
            \For{$\epsilon \in \{+, 0, -\}$}{
            $v = \bar{{v}}_{j,k}^{\epsilon} $\;\\
            {$\pi(i, v) = \max_{u \in \mathcal{T}} \{ \pi(i-1, u) + \psi_{\bar{u}, \bar{v}}  +  f_t(\vec{h}_i)_{\bar{\vec{v}}} + f_s([\vec{g}_{i+j,i+k};\overleftarrow{\vec{h}_{i}}])_{\epsilon} + f_o(\vec{g}_{i+j,i+k})  + f_r(j,k) \}$}}}}}
}

Backward for the optimal sequence $\vec{y}^{*} = \{ \vec{y}^{*}_1 \cdots \vec{y}^{*}_n \}$\;
\For{$i \in \{n, \cdots ,1 \}$} {
\If{$i=n$}{$\vec{y}^{*}_n = \argmax_{v \in \mathcal{T}} {\{ \pi(n,v) + \psi_{\bar{v}, STOP} \}}$}
\Else{$\vec{y}^{*}_i = \argmax_{v \in \mathcal{T}} {\{ \pi(i,v) + \psi_{\bar{v}, \bar{\vec{y}}^{*}_{i+1}} \}}$}
}
\caption{Decoding based on Viterbi}
\end{algorithm}

\end{document}